\pdfoutput=1
\documentclass[11pt]{article}
\usepackage[preprint]{acl}
\usepackage{frontpage}

\usepackage{times}
\usepackage{latexsym}
\usepackage[T1]{fontenc}
\usepackage[utf8]{inputenc}
\usepackage{microtype}
\usepackage{inconsolata}
\usepackage{graphicx}
\usepackage{amsmath,amssymb}
\usepackage{xcolor}
\usepackage{caption}
\usepackage{hyperref}
\usepackage[most]{tcolorbox}

\definecolor{plotblue}{HTML}{4C72B0}
\definecolor{plotgreen}{HTML}{55A868}
\definecolor{plotred}{HTML}{C44E52}
\definecolor{qacolor}{RGB}{45,70,140}
\definecolor{subjcolor}{RGB}{46,125,70}
\definecolor{dircolor}{RGB}{192,96,40}
\hypersetup{colorlinks=true, linkcolor=qacolor, citecolor=qacolor, urlcolor=qacolor}

\newcommand{\routetitle}[2]{\texorpdfstring{\textcolor{#1}{#2}}{#2}}
\newcommand{\subjroute}[1][subject-mediated route]{\textcolor{subjcolor}{\emph{#1}}}
\newcommand{\dirroute}[1][direct route]{\textcolor{dircolor}{\emph{#1}}}
\newcommand{\takeaway}[1]{\textbf{#1}}
\newcommand{\boxlabel}[1]{{\setlength{\fboxsep}{1.2pt}\setlength{\fboxrule}{0.4pt}\fbox{#1}}}
\newcommand{\qaA}{{\color{qacolor!70!black}\bfseries A:~}}
\newtcolorbox{qabox}{
  enhanced, breakable,
  colback=qacolor!4, frame hidden,
  borderline west={3pt}{0pt}{qacolor},
  left=7pt, right=6pt, top=5pt, bottom=5pt,
  before skip=8pt, after skip=10pt,
  fontupper=\small
}

\title{Tracing How Language Models Integrate Temporal Anchors into Factual Recall}
\fpauthors{Ayana Niwa \quad Masahiro Kaneko \quad Kentaro Inui}
\fpaffil{\includegraphics[height=1.15em]{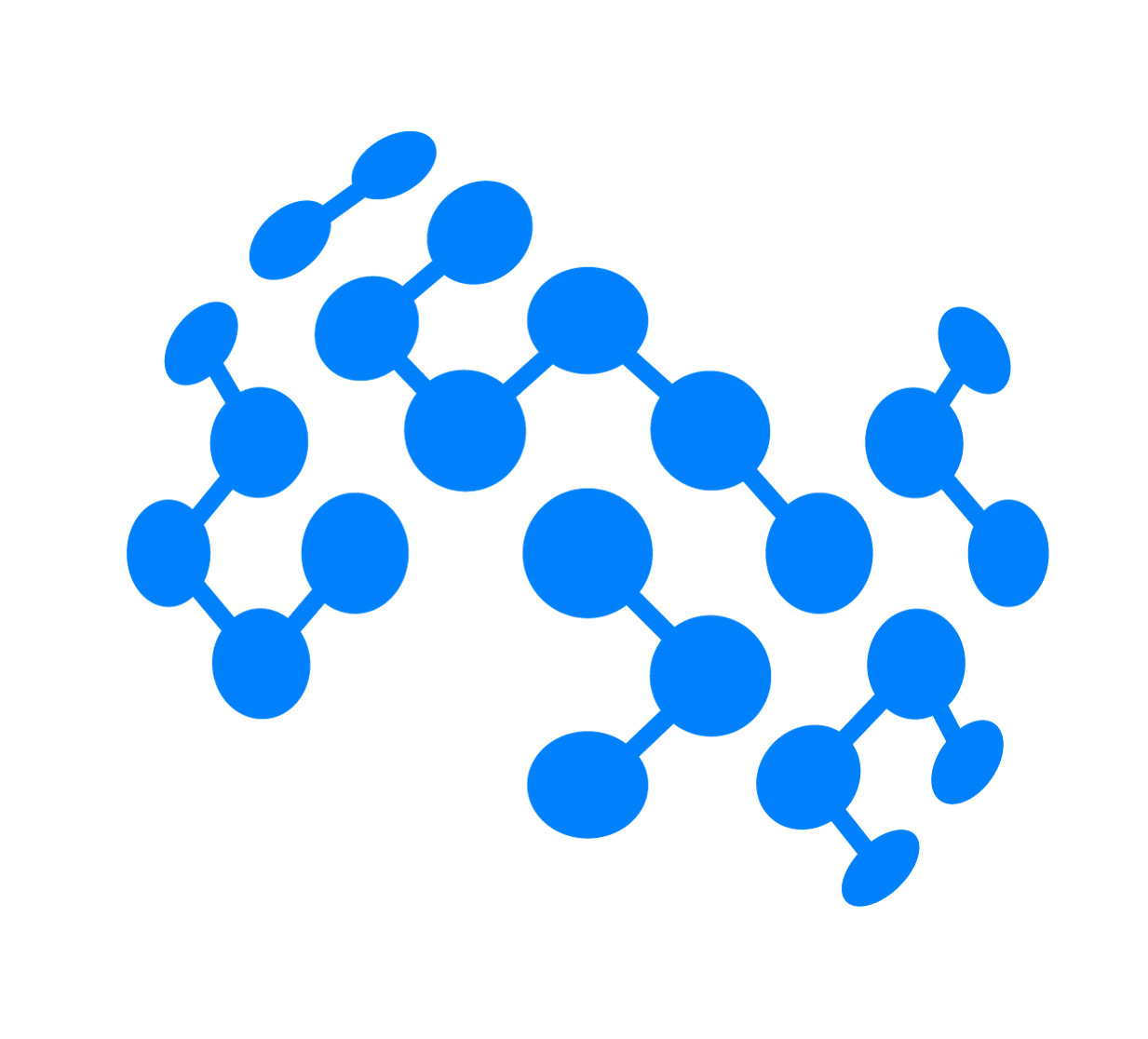}}{Mohamed bin Zayed University of Artificial Intelligence}
\fpemails{\{ayana.niwa, masahiro.kaneko, kentaro.inui\}@mbzuai.ac.ae}

\begin{document}
\maketitle

\begin{abstract}
  Factual recall in language models is typically framed as retrieving a factual association: a subject--relation pair maps to a single object.
  However, many facts are context-dependent: the same pair maps to different objects depending on a condition such as time.
  We introduce \textbf{\emph{anchored factual recall}}, in which the model receives an explicit conditioning expression, an \emph{anchor}, and must select the object valid under it.
  This raises questions that single-object recall does not: where the anchor enters the computation, what it changes, and how that change selects the answer.
  Focusing on temporal anchors, we answer them with causal interventions on four open-weight models on English prompts and identify two routes.
  In a \subjroute[subject-mediated route], the anchor resolves the final subject-token representation to the queried year, and the relation redistributes the readout of this state across attention heads.
  In a \dirroute[direct route], the final position attends to the anchor, and upper-layer MLPs turn this signal into a preference for the time-appropriate object.
  The two routes are parallel channels: their contributions add up, most queries use both, and their balance depends on the query, e.g., on the relation, consistently across models.
  This reframes mechanistic accounts of factual recall from retrieving a single stored object to selecting among context-dependent alternatives.
\end{abstract}

\section{Introduction}
\label{sec:intro}

Language models (LMs) encode large amounts of factual knowledge in their parameters, and a growing body of work characterizes how this knowledge is retrieved during inference~\citep{petroni2019language,roberts2020much,dai2022knowledge,niwa2025rectifying}.
An influential account by \citet{geva-etal-2023-dissecting} describes recall for prompts such as ``\textit{The capital of France is \_\_}'' in three stages: the model enriches the subject representation with attributes, propagates relation information to the final position, and extracts the target attribute there. Related studies support key parts of this picture~\citep{meng2022locating,yu2023characterizing,hernandez2024linearity}.

This account addresses facts represented as (\texttt{subject}, \texttt{relation}, \texttt{object}) triples, in which the subject and relation determine a single object, so recall amounts to retrieving that object.
However, many real-world facts pose a different problem: ``\textit{Cristiano Ronaldo played for \_\_}'' has no unique answer until ``\textit{in 2010}'' or ``\textit{in 2020}'' resolves it to ``\textit{Real Madrid}'' or ``\textit{Juventus}.''
Here several objects are valid for the same subject and relation, and the model must select among them according to a condition.
Selection raises questions unique to it, which do not arise in the recall of triple-form facts: \emph{where} the condition enters the computation (Q1), \emph{what} it changes (Q2), and \emph{how} that change makes one object preferred over its competitors (Q3).

\begin{figure}[t]
  \centering
  \includegraphics[width=\linewidth]{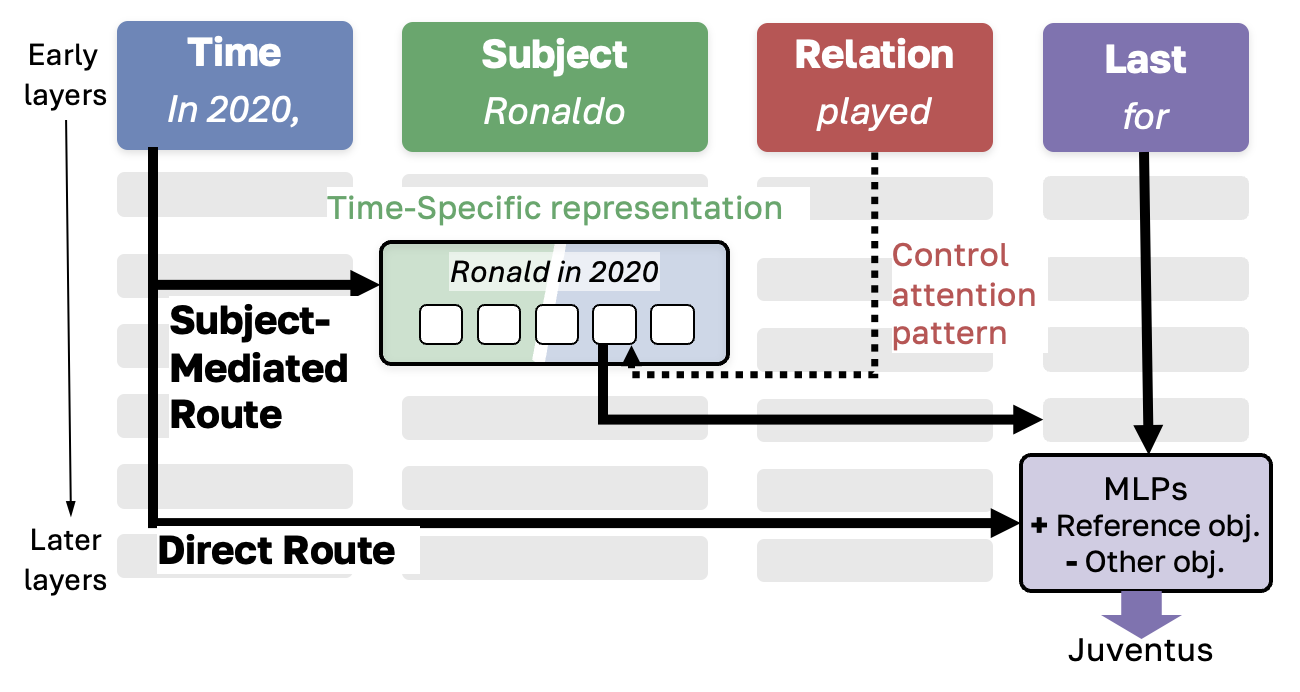}
  \caption{\takeaway{Two routes integrate the temporal anchor into factual recall.} \subjroute[Subject-mediated]: the anchor resolves the subject representation to 2020, and the relation sets how the final position reads it. \dirroute[Direct]: the final position reads the anchor itself, and MLPs turn it into a preference for \textit{Juventus}. The two routes are used jointly on most queries, with a balance that varies across queries.}
  \label{fig:outline}
\end{figure}

We study \textbf{\textit{anchored factual recall}}, in which the model receives an explicit conditioning expression, an \textit{anchor}, alongside the subject and relation: given (\underline{\texttt{anchor}}, \texttt{subject}, \texttt{relation}), it must recall the \texttt{object}.
We focus on \textbf{temporal anchors}, a central case in which the correct object changes over time~\citep{tan-etal-2023-towards,zhao-etal-2024-set}.
Work on time-conditioned recall has almost exclusively examined model outputs~\citep{dhingra2022time,chen2021a,herel2025timeawarenesslargelanguage}.
\citet{park-etal-2025-time} identify attention heads specialized for temporal knowledge, but how the anchor's information is routed and transformed to select one object over its competitors remains untraced.

We answer Q1--Q3 with causal interventions across four models on English prompts, and find the following (\autoref{fig:outline}).
\textbf{Q1} (\emph{where}): the anchor enters at two points: the subject span uses it in early-to-middle layers, and the final position accesses it directly from the middle layers onward, defining a \subjroute{} and a \dirroute.
\textbf{Q2} (\emph{what}): the final subject-token state becomes specific to the queried year and selectively supports the object valid at that year.
\textbf{Q3} (\emph{how}): on the subject-mediated route, the relation sets which attention heads read the subject state out, and these heads causally support the prediction.
On the direct route, the final position reads the anchor, and upper-layer MLPs convert this signal into a preference for the time-appropriate object.
The two routes are parallel channels: their contributions add up, most queries use both, and the balance between them varies with the query, consistently across models.

Together, these results reframe the standard picture of factual recall from \emph{retrieving} an object to \emph{selecting} one among competing objects, a selection that the model carries out through two parallel, jointly used routes.
To our knowledge, this is the first account of the information flow by which a temporal anchor selects among the competing facts an LM stores: where its information enters, how it is routed and transformed, and how the two routes combine.
More broadly, it provides a starting point for understanding how LMs recall knowledge whose validity depends on the conditions under which it is queried.

\section{Task Definition}
\label{sec:background-questions}

\subsection{Anchored Factual Recall}

Mechanistic studies of factual recall typically consider facts represented as triples $(s,r,o)$, in which the object $o$ is treated as determined by the subject $s$ and relation $r$.
This formulation supports the study of how an LM retrieves a fixed factual association from a subject--relation prompt, but it does not represent cases in which the valid object depends on an additional condition.

We instead consider \emph{anchored facts}, represented as quadruples $(a,s,r,o)$, where an anchor $a$ conditions which object is valid.
For a fixed subject--relation pair, different anchors may select different objects.
Writing $f$ for the map from a query $(a,s,r)$ to its gold object, this means $f(a,s,r) \neq f(a',s,r)$ for some $a \neq a'$.
In \textbf{temporally anchored factual recall}, the anchor is a year, and the model must select the object valid at that time.
Anchored factual recall therefore poses a different mechanistic problem from fixed-association recall: the model must not only retrieve factual information, but also use the anchor to resolve competition among context-dependent alternatives.

Prior work on unanchored factual recall provides causal tools for tracing how internal components contribute to a prediction~\citep{geva-etal-2023-dissecting,meng2022locating,yu2023characterizing,hernandez2024linearity}.
We build on this intervention-based methodology, but do not assume that a computational decomposition developed for subject--relation prompts applies to anchored factual recall.

\subsection{Scope and Notation}
Anchored factual recall is a general formulation; we study its temporal case, in which the anchor is a year and the model must select the object valid at that time.
Because we ask how the anchor selects the object, we need queries in which the object is predicted last and only the anchor varies.
We thus use fixed English anchor-first queries,
\[
  \texttt{In [year], [subject] [relation]}.
\]
Our findings are scoped to this setting.

Extending the positional decomposition of \citet{geva-etal-2023-dissecting} with an anchor, we partition each tokenized query into four spans: the \textit{anchor span} $\mathcal{A}$; the \textit{subject span} $\mathcal{S}$; the \textit{last-position span} $\mathcal{L}$, the final query token alone; and the \textit{relation span} $\mathcal{R}$, the relation tokens preceding it.
For example, for the query ``\textit{In 2010, Cristiano Ronaldo played for}'', $\mathcal{A}$ covers ``\textit{2010}'', $\mathcal{S}$ covers ``\textit{Cristiano Ronaldo}'', $\mathcal{R}$ covers ``\textit{played}'', and $\mathcal{L}$ is ``\textit{for}''; the remaining tokens (``\textit{In}'' and the comma) belong to no span.
Thus, $\mathcal{R}\cup\mathcal{L}$ realizes the full relation expression, and the spans occur in the fixed order $\mathcal{A} \prec \mathcal{S} \prec \mathcal{R} \prec \mathcal{L}$.
Following the same source--target convention, we write $\mathcal{Y}\!\leftarrow\!\mathcal{X}$ to denote attention from query positions in the target span $\mathcal{Y}$ to key--value positions in the source span $\mathcal{X}$.
For example, $\mathcal{L}\!\leftarrow\!\mathcal{S}$ denotes attention from the final position to the subject span, whereas $\mathcal{R}\!\leftarrow\!\mathcal{A}$ denotes attention from the relation span to the anchor span.

\begin{figure*}[t]
  \centering
  \includegraphics[width=\linewidth]{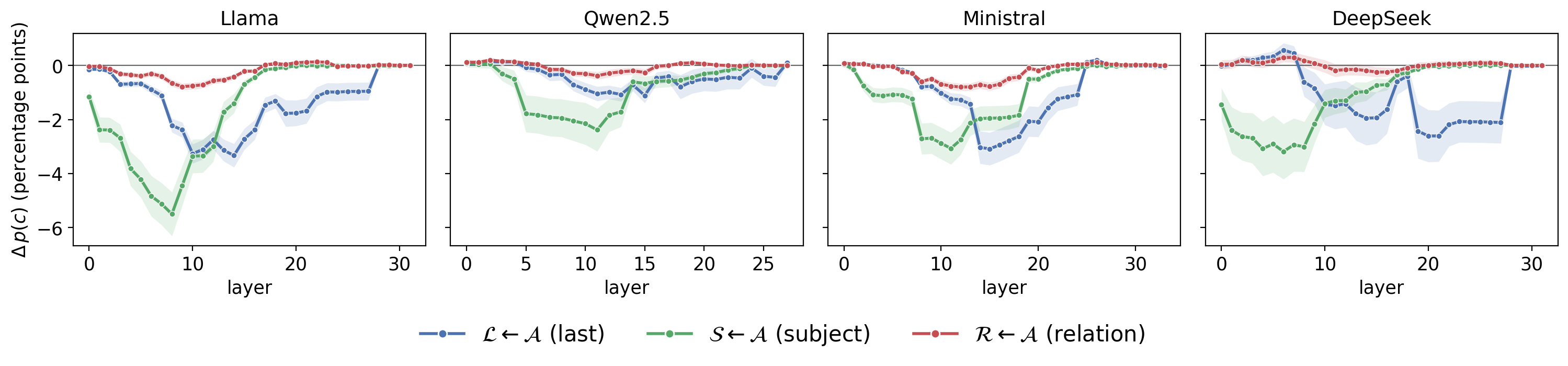}
  \caption{\takeaway{Q1: the anchor enters the computation at two points.} Attention knockout of \textcolor{plotgreen}{$\mathcal{S}\!\leftarrow\!\mathcal{A}$} degrades the prediction in early-to-middle layers, knockout of \textcolor{plotblue}{$\mathcal{L}\!\leftarrow\!\mathcal{A}$} from the middle layers onward, whereas knockout of \textcolor{plotred}{$\mathcal{R}\!\leftarrow\!\mathcal{A}$} has little effect.
  }
  \label{fig:attention_ko}
\end{figure*}
\section{Experimental Setup}
\label{sec:method}

\paragraph{Data.}
We use TempLAMA~\citep{dhingra2022time}\footnote{
Analyzing a single controlled dataset in depth is common in mechanistic studies, e.g., the IOI template~\citep{wang2023interpretability} and CounterFact~\citep{geva-etal-2023-dissecting}.}, a dataset of temporally scoped factual queries built from Wikidata, in which each subject--relation template is paired with a year between 2010 and 2020 and the gold object may vary across years.
It represents time-scoped facts as quadruples $(a,s,r,o)$: in our setting, the anchor $a$ is a year, and the valid object $o$ may change across years for a fixed subject--relation pair $(s,r)$.
We group instances by subject--relation pair and restrict the analysis to groups for which the gold object changes across years and the model correctly predicts the gold object for at least two years whose gold objects differ.
For every retained group, we include all year-specific instances the model answers correctly.
This restriction focuses our analysis on cases in which the model demonstrably exhibits time-sensitive factual recall.\footnote{Restricting mechanistic analysis to correctly answered instances follows standard practice in studies of factual recall~\citep{geva-etal-2023-dissecting,meng2022locating}.}
We count a prediction as correct if its greedy continuation contains the canonical gold object or its accepted aliases under a normalized, word-bounded matching criterion.
We keep only years with a single gold object.

\paragraph{Prompts.}

Each prompt consists of five in-context demonstrations followed by the query defined in \S\ref{sec:background-questions}; each demonstration is a completed query of the same form, ending with its gold object.
For each retained instance, we randomly sample five demonstrations that share neither year, subject, nor relation with it, so that no demonstration leaks the answer or its query pattern.
The sampled demonstrations are fixed for that instance and reused across all analyses of that instance and all year-swapped contrast runs derived from it.

\paragraph{Models.}

We benchmarked several open-weight 7--8B models on TempLAMA and retained the four highest-scoring, ensuring that enough instances survive the filter above: Llama-3.1-8B~\citep{grattafiori2024llama3herdmodels}, Qwen2.5-7B~\citep{qwen2.5}, Ministral-8B~\citep{liu2026ministral3}\footnote{\texttt{Ministral-3-8B-Base-2512}}, and DeepSeek-R1-Distill-Llama-8B~\citep{deepseekai2025deepseekr1incentivizingreasoningcapability}.

\paragraph{Reference--contrast pairs.}

To isolate the anchor's effect, most interventions compare a query with a copy differing only in the year.
For each instance, which we call the \emph{reference}, we sample one \emph{contrast} year from the same group whose gold object differs from the reference's, and replace only the year in the query; demonstrations and all other tokens are unchanged.
We keep a pair only if the model answers both prompts correctly under these shared demonstrations.
We call the two forward passes the \emph{reference run} and the \emph{contrast run}, and write $c$ and $\bar{c}$ for the first tokens of the canonical reference and contrast objects; pairs in which the two share the first token are excluded.
All interventions apply to the query segment only; their layer scope is given in Appendix~\ref{app:impl}.

\section{Mechanistic Analysis}\begin{figure*}
  \centering
  \includegraphics[width=\linewidth]{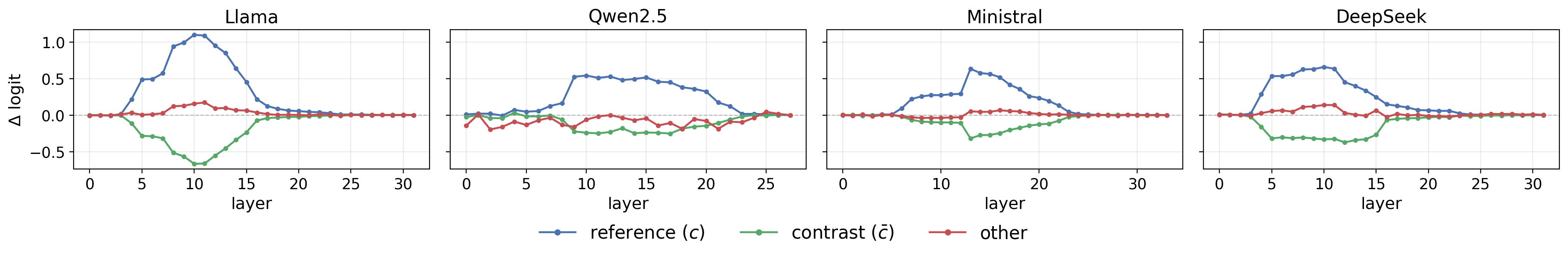}
  \caption{ \takeaway{The subject state selectively supports the year-appropriate object.} Patching the reference subject state into the contrast run raises the logit of the reference-year object $c$, lowers the logit of the contrast-year object $\bar{c}$, and leaves the mean logit of the other temporal candidates largely unchanged. }
  \label{fig:single_year}
\end{figure*}

\label{sec:analysis}

Our analysis follows the three questions from \S\ref{sec:intro}: \textit{where} the model uses the anchor (\textbf{Q1}; \S\ref{sec:flow}), \textit{what} the anchor changes (\textbf{Q2}; \S\ref{sec:subject}), and \textit{how} that change is turned into a preference for the time-appropriate object (\textbf{Q3}; \S\ref{sec:direct}).

\subsection{Q1: \textit{Where} Does the Anchor Enter?}
\label{sec:flow}

\begin{qabox}
  \qaA Temporal information enters through two routes: a \subjroute{}
  (subject span, early-to-middle layers) and a \dirroute{} (final position, middle layers onward).
\end{qabox}

\paragraph{\boxlabel{Method.}}
We first ask at which positions, and at which layers, the anchor's information enters the recall computation.
An entry point is a position whose attention to the anchor the prediction depends on; to find these, we follow \citet{geva-etal-2023-dissecting} and apply \emph{attention knockout} to each edge $\mathcal{Y}\!\leftarrow\!\mathcal{A}$ for $\mathcal{Y}\in\{\mathcal{S},\mathcal{R},\mathcal{L}\}$: over a window of adjacent layers centered at $\ell$, we block all heads at $\mathcal{Y}$ from attending to the anchor span and measure the change in the reference-token probability,
\begin{equation}
  \Delta p_{\mathcal{Y}\leftarrow\mathcal{A}}(\ell) = p_{\mathrm{KO}(\mathcal{Y}\leftarrow\mathcal{A},\ell)}(c) - p_{\mathrm{base}}(c),
\end{equation}
where $p_{\mathrm{KO}(\mathcal{Y}\leftarrow\mathcal{A},\ell)}$ and $p_{\mathrm{base}}$ are the next-token probabilities of $c$ at the final position in the knockout run and in the unmodified run of the same prompt.
A negative $\Delta p$ indicates that attention from $\mathcal{Y}$ to the anchor causally supports the reference prediction in the tested layer window.

\paragraph{\boxlabel{Result.}}
Fig.~\ref{fig:attention_ko} shows $\Delta p$ for the three edges.
Across all four models, blocking $\mathcal{S}\!\leftarrow\!\mathcal{A}$ produces the earliest substantial decrease in the reference-token probability, with the effect concentrated in the early-to-middle layers.
Blocking $\mathcal{L}\!\leftarrow\!\mathcal{A}$ has little effect in the earliest layers but produces a later decrease from the middle of the network onward.
By contrast, blocking $\mathcal{R}\!\leftarrow\!\mathcal{A}$ has a much smaller effect: a shallow dip in the middle layers.
Although the magnitude and layer range of each effect vary across models, this layer-wise ordering is consistent across the models.

\textbf{Temporal information thus becomes causally relevant at two points in the recall computation.}
At the subject span, where blocking $\mathcal{S}\!\leftarrow\!\mathcal{A}$ degrades the prediction early, the subject tokens read the anchor; at the final position, where blocking $\mathcal{L}\!\leftarrow\!\mathcal{A}$ degrades it later, the anchor is read directly, without passing through the subject.
The relation span, by contrast, shows little sign of being a major carrier of temporal information: blocking $\mathcal{R}\!\leftarrow\!\mathcal{A}$ costs little.
One possible reason is that the subject and the anchor together already constrain the relation to the subject's time-varying attributes, leaving the relation span less to gain from reading the anchor.
We call the paths through the subject span and through the final position the \subjroute{} and the \dirroute.
Here \emph{route} means a path by which anchor information reaches the prediction, not a separate set of components.
\S\ref{sec:subject} and \S\ref{sec:relation-readout} characterize the first, \S\ref{sec:direct-mlp} the second, and \S\ref{sec:two_routes_combination} how they relate.

\subsection{Q2: \textit{What} Does the Anchor Change?}
\label{sec:subject}

\begin{qabox}
  \qaA The anchor changes the final subject-token representation into a largely entity-specific state that causally and selectively supports the year-appropriate object.
\end{qabox}

\paragraph{\boxlabel{Method.}}

We apply activation patching~\citep{meng2022locating,wang2023interpretability}: at each layer $\ell$, we replace the residual-stream state at the final subject token, where subject information accumulates~\citep{geva-etal-2023-dissecting,meng2022locating}, in the contrast run with that of the reference run.
Because the two prompts differ only in the anchor, any change in the prediction reflects anchor-dependent information in the patched state.
We measure the logits of the first continuation tokens of three classes of temporal candidates: the reference object $c$, the contrast object $\bar c$, and \emph{other}, the mean over the group's remaining gold objects (computed for groups that have at least one besides $c$ and $\bar c$).
We report the change in each logit relative to the unmodified contrast run, so that zero corresponds to no effect of the patch.
If the subject state did not depend on the anchor, patching would change none of the three logits.
If it is specific to the reference anchor, it should raise $c$, lower $\bar c$, and leave the \emph{other} candidates unchanged: the first two show that the patch replaces one year's preference with another's, and the third that this replacement does not spill over to the objects of other years.

\paragraph{\boxlabel{Result.}}
Across all four models, patching the reference subject state into the contrast run selectively shifts the prediction toward the reference-year object (Fig.~\ref{fig:single_year}).
The logit of $c$ increases, the logit of $\bar c$ decreases, and the mean logit of the remaining temporal candidates changes little.
The effect emerges and reaches its maximum in the middle layers.
Rather than broadly increasing all objects associated with the pair, the intervention selectively favors the reference-year object, suggesting that the patched representation carries anchor-dependent information that downstream computation uses to distinguish among temporal alternatives.

One alternative explanation is that the anchor simply adds a generic year-dependent component to the subject state, shared across entities. We test this possibility in Appendix~\ref{sec:subject-additive-null} by estimating such a shared year offset from other subjects. Across all four models, this offset recovers at most about a quarter of the effect produced by the true reference-year subject state. The anchor-dependent change is therefore largely specific to the queried subject, rather than a generic representation of the year. Taken together with the selective patching effect above, this suggests that \textbf{the anchor resolves the subject representation toward the object that is valid for that subject at the queried time.}

\begin{figure*}[t]
  \centering
  \includegraphics[width=\linewidth]{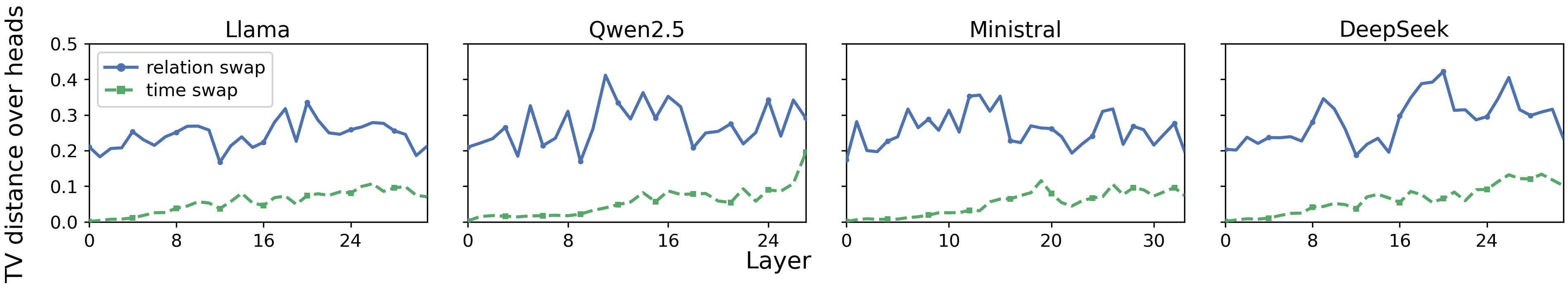}
  \caption{
    \takeaway{The relation redistributes last-subject readout across attention heads.}
    Per-layer total-variation distance between head-normalized last-subject attention distributions (Eq.~\ref{eq:headwise-readout-tv}): swapping the relation changes the distribution far more than swapping the anchor year.
  }
  \label{fig:relation-readout-distribution}
\end{figure*}
\subsection{Q3: \textit{How} Is the Answer Selected?}
\label{sec:direct}

\begin{qabox}
  \qaA Through both routes: on the \subjroute[subject-mediated route], the relation redistributes readout of the time-conditioned subject state across attention heads, and relation-sensitive heads causally support the prediction (\S\ref{sec:relation-readout}); on the \dirroute[direct route], upper-layer MLPs turn the anchor signal into a preference for the year-appropriate object (\S\ref{sec:direct-mlp}).
  The two routes are parallel channels, jointly used on most queries, and the balance between them varies with the query (e.g., the relation), consistently across models (\S\ref{sec:two_routes_combination}).
\end{qabox}

\subsubsection{
\routetitle{subjcolor}{Subject-Mediated Route}: The Relation Reconfigures the Head-Wise Readout from the Subject State}
\label{sec:relation-readout}

We begin with the \subjroute.
The previous section showed that the anchor makes the subject state selective for the object valid at the queried time.
The final position can use this state only by attending to it, so we next ask how the relation controls that attention: which heads read from the subject, and whether they matter.

\paragraph{\boxlabel{Method.}}
Let $\alpha_{\ell h}(x)$ denote the attention weight from the final position $\mathcal{L}$ to the final subject token $s^\star$ in head $h$ of layer $\ell$ for prompt $x$.
We compare two swaps while holding the demonstrations fixed.
A \emph{relation swap} replaces the relation expression $\mathcal{R}\cup\mathcal{L}$ with another relation having the same number of tokens, so that token positions remain aligned.
A \emph{time swap} changes only the anchor year.

\underline{\emph{(i) Redistribution across heads.}}
We ask whether changing the relation changes \emph{which heads} read from the subject.
Within each layer, we normalize the last-subject attention weights across heads,
\[
  \pi_{\ell h}(x)
  =
  \frac{\alpha_{\ell h}(x)}
  {\sum_j \alpha_{\ell j}(x)},
\]
and measure the total-variation distance between the resulting distributions:
\begin{equation}
  D_{\mathrm{head}}^{(\ell)}(x,x')
  =
  \frac{1}{2}
  \sum_h
  \left|
  \pi_{\ell h}(x)-\pi_{\ell h}(x')
  \right|.
  \label{eq:headwise-readout-tv}
\end{equation}
This normalization removes changes in the total attention paid to $s^\star$ within a layer and isolates changes in how that readout is distributed across heads.
A larger distance under the relation swap than the time swap indicates that the head-wise allocation of subject readout depends on the queried relation rather than merely on the temporal anchor.

\begin{figure}[t]
  \centering
  \includegraphics[width=\linewidth]{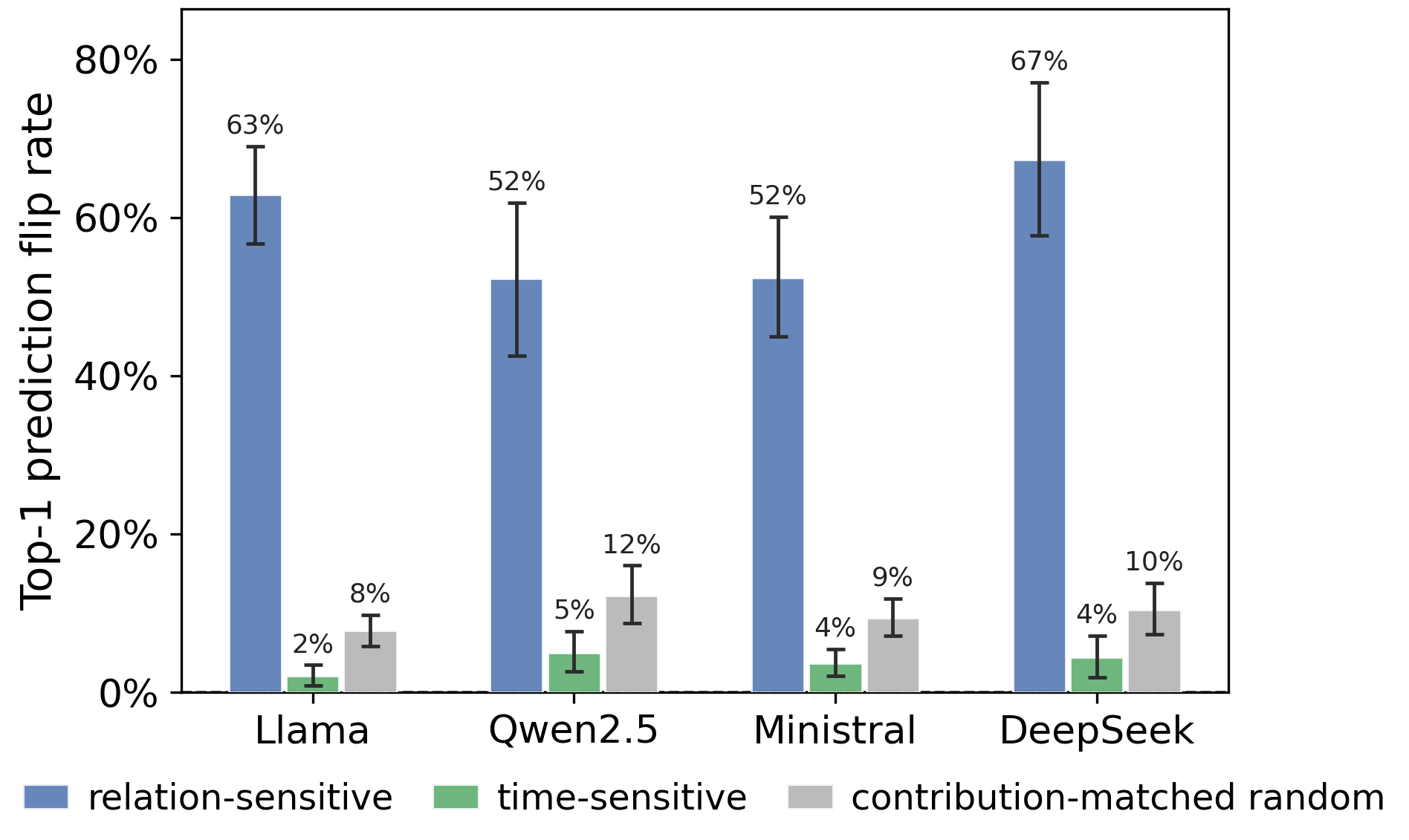}
  \caption{
    \takeaway{Relation-sensitive heads causally support the correct prediction.}
    Top-1 flip rate after ablating the last-subject contribution of relation-sensitive, time-sensitive, or contribution-matched random heads (selected on discovery folds; \S\ref{sec:relation-readout}).
    Error bars: 95\% bootstrap CIs.
  }
  \label{fig:relation-head-ablation}
\end{figure}
\underline{\emph{(ii) Cross-fitted ablation.}}
A change in attention weights is descriptive: a head can change its readout without affecting the answer.
We therefore ask whether heads whose readout from the subject is particularly sensitive to the relation are causally important for the prediction.

We split subject--relation groups into five folds.
On the discovery folds, each layer--head pair receives a relation-sensitivity score
\[
  s_{\ell h}
  =
  \mathbb{E}_{\mathrm{disc}}
  \left[
    \left|
    \alpha_{\ell h}^{\mathrm{base}}
    -
    \alpha_{\ell h}^{\mathrm{rel}}
    \right|
    -
    \left|
    \alpha_{\ell h}^{\mathrm{base}}
    -
    \alpha_{\ell h}^{\mathrm{time}}
    \right|
  \right],
\]
where \emph{base}, \emph{rel}, and \emph{time} denote the original, relation-swapped, and time-swapped prompts.
The score is high when a head's attention to $s^\star$ changes more when the relation is swapped than when only the anchor year is swapped.
Unlike the redistribution analysis above, this score uses the unnormalized weights: causal relevance may depend not only on how subject readout is redistributed across heads, but also on how much a head reads from the subject in absolute terms.
We select the top $10\%$ of layer--head pairs without using answer logits.

We then evaluate the selected pairs on held-out correctly answered base prompts.
Instead of ablating a whole head, we remove its contribution from the final subject token $s^\star$ to the final position $\mathcal{L}$:
\begin{equation}
  \mathbf{c}_{\ell h}
  =
  \alpha_{\ell h}\,
  W_O^{(\ell,h)}
  \mathbf{V}^{(\ell,h)}(s^\star),
  \label{eq:relation-head-contribution}
\end{equation}
where $W_O^{(\ell,h)}$ is the head's output projection and $\mathbf{V}^{(\ell,h)}(s^\star)$ its value vector at $s^\star$.
This contribution is recomputed during the intervened forward pass, so ablations in later layers operate on the state produced by earlier interventions.
We report the top-1 flip rate, the fraction of prompts for which the originally correct top-ranked next token is no longer top-ranked after ablation.
As controls, we ablate \emph{time-sensitive} layer--head pairs selected by the reversed score (time swap minus relation swap) and \emph{contribution-matched random} pairs from the same layers (Appendix~\ref{app:impl}).

\paragraph{\boxlabel{Result.}}
\underline{\emph{(i) Redistribution.}}
Changing the relation alters the head-wise distribution of last-subject readout far more than changing only the anchor year (Fig.~\ref{fig:relation-readout-distribution}).
Across all four models, the distance under the relation swap exceeds that under the time swap at every layer.
\textbf{Changing the relation thus redistributes the readout of the time-conditioned subject state across attention heads.}

\underline{\emph{(ii) Ablation.}}
The cross-fitted ablation shows that \textbf{relation-sensitive subject readout is causally important for the answer} (Fig.~\ref{fig:relation-head-ablation}).
Removing the last-subject contribution of the selected layer--head pairs changes the originally correct top-1 next-token prediction in approximately $63\%$ of Llama prompts, $52\%$ of Qwen prompts, $52\%$ of Ministral prompts, and $67\%$ of DeepSeek prompts.
By contrast, ablating time-sensitive pairs changes the prediction in only $2$--$5\%$ of prompts, and contribution-matched random pairs in $8$--$12\%$.
For every model, the paired difference between relation-sensitive and contribution-matched random ablation is positive.

With \S\ref{sec:subject}, these results support a division of labor along the subject-mediated route: the anchor makes the subject state selective for the object valid at that time, the relation determines how that state is read out at the final position, and relation-sensitive heads causally support the prediction.

\subsubsection{\routetitle{dircolor}{Direct Route}: Final-Position MLPs Convert the Anchor Signal into an Answer Preference}
\label{sec:direct-mlp}

We now turn to the \dirroute.
The knockout analysis showed that the final position $\mathcal{L}$ reads directly from the anchor, but not how the resulting signal becomes selective for the correct temporal object.
We therefore ask two questions: whether final-position MLPs are necessary for the answer-relevant effect of this direct signal (i), and whether their outputs become aligned with the reference-over-contrast answer direction (ii).

\paragraph{\boxlabel{Method.}}
\underline{\emph{(i) Value injection with MLP pinning.}}
We first transfer only the anchor-derived attention signal from the reference run to the contrast run.
For each layer $\ell$, let $\mathbf{a}_{\mathcal{A}}^{(\ell)}(x)$ denote the contribution of the anchor positions to the attention output at $\mathcal{L}$ for prompt $x$.
We compute
\[
  \Delta \mathbf{a}_{\mathcal{A}}^{(\ell)}
  =
  \mathbf{a}_{\mathcal{A}}^{(\ell)}(x_{\mathrm{ref}})
  -
  \mathbf{a}_{\mathcal{A}}^{(\ell)}(x_{\mathrm{ctr}}),
\]
where $x_{\mathrm{ref}}$ and $x_{\mathrm{ctr}}$ are the reference and contrast prompts, and add this difference to the attention output at $\mathcal{L}$ in the contrast run at every layer simultaneously.
Thus, the contrast run receives the reference--contrast difference carried specifically along $\mathcal{L}\!\leftarrow\!\mathcal{A}$, while the rest of the computation begins from the contrast prompt.

We quantify how much this intervention restores the reference answer using the logit difference
\[
  \mathrm{LD}=z_c-z_{\bar{c}},
\]
where $z_v$ is the final-position logit of token $v$, and the recovery score
\begin{equation}
  R
  =
  \frac{
    \mathrm{LD}_{\mathrm{after}}-\mathrm{LD}_{\mathrm{ctr}}
  }{
    \mathrm{LD}_{\mathrm{ref}}-\mathrm{LD}_{\mathrm{ctr}}
  }.
  \label{eq:recovery}
\end{equation}
Here, $R=0$ corresponds to the unmodified contrast run and $R=1$ to the reference-run value; we leave the score unclipped.
Injection with the MLPs free to respond gives $R_{\mathrm{total}}$.

To test whether final-position MLPs are responsible for this recovery, we repeat the same injection while fixing every final-position MLP output to the value it had in the unmodified contrast run.
The remaining recovery is $R_{\mathrm{non\text{-}MLP}}$.
We define
\[
  R_{\mathrm{MLP}}
  =
  R_{\mathrm{total}}
  -
  R_{\mathrm{non\text{-}MLP}},
\]
the additional recovery when the final-position MLPs are allowed to respond to the injected anchor signal.
Each reference--contrast pair therefore yields one value of $R_{\mathrm{total}}$, $R_{\mathrm{non\text{-}MLP}}$, and $R_{\mathrm{MLP}}$.
If the final-position MLPs mediate the effect of the direct anchor signal, pinning them should remove the recovery, so that $R_{\mathrm{non\text{-}MLP}}$ is near zero and $R_{\mathrm{MLP}}$ accounts for most of $R_{\mathrm{total}}$; if they do not, $R_{\mathrm{non\text{-}MLP}}$ should remain close to $R_{\mathrm{total}}$.

\begin{figure}
  \centering
  \includegraphics[width=\linewidth]{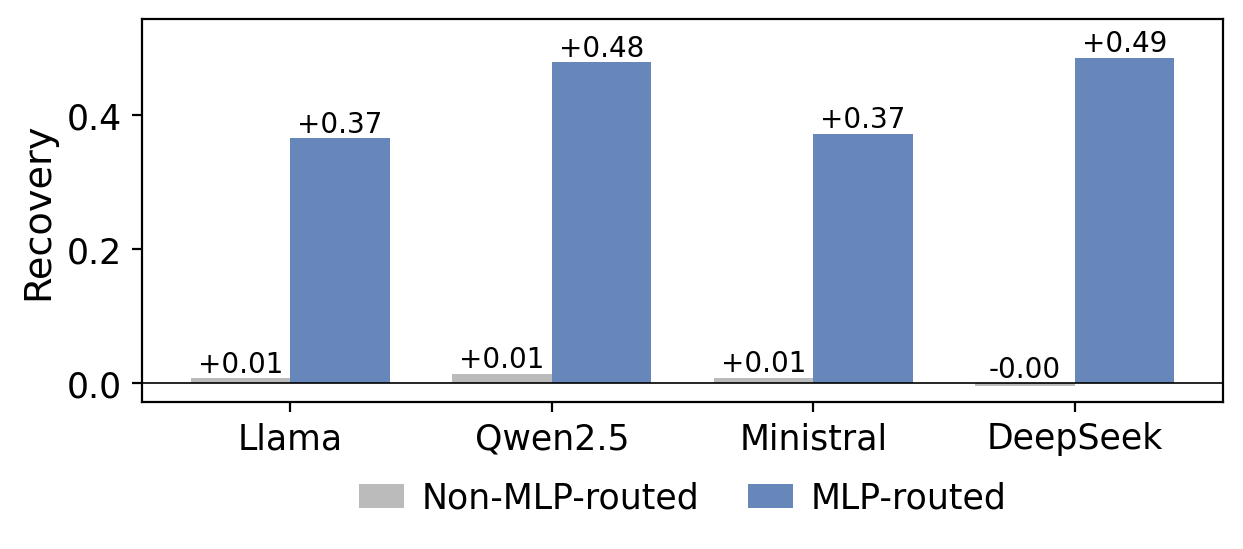}
  \caption{\takeaway{The direct route acts through final-position MLPs.}
  Value-injection recovery decomposed into MLP-mediated ($R_{\mathrm{MLP}}$) and non-MLP ($R_{\mathrm{non\text{-}MLP}}$) components: pinning the MLPs removes nearly all of the recovery.}
  \label{fig:mlp_routed}
\end{figure}
\begin{figure}
  \centering
  \includegraphics[width=0.7\linewidth]{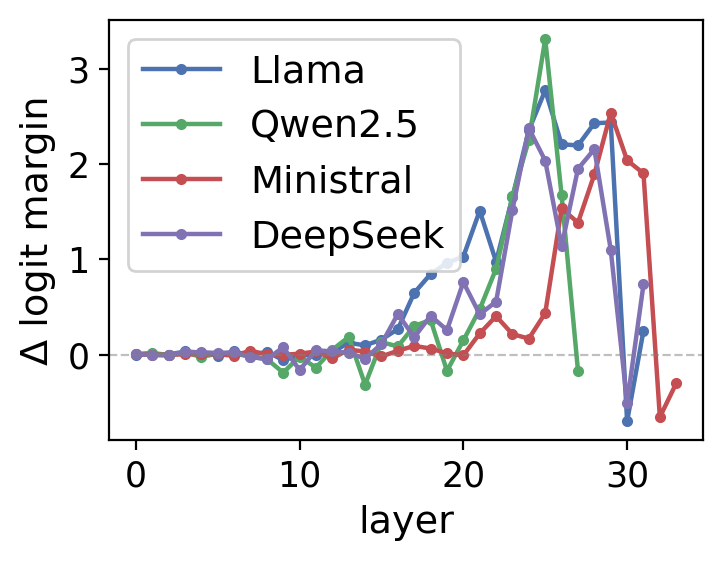}
  \caption{\takeaway{Upper-layer MLPs are answer-selective.}
  Direct logit attribution: the reference-minus-contrast difference $\Delta\mathrm{DLA}^{(\ell)}$ in each final-position MLP's contribution to the logit margin turns positive over a band of upper layers.}
  \label{fig:mlp_readout}
\end{figure}
\underline{\emph{(ii) Direct logit attribution (DLA)}}~\citep{nostalgebraist2020logitlens}.
The intervention above tests whether the MLPs mediate the effect of the direct anchor signal.
We next ask what answer preference their outputs write into the residual stream.
For each layer, we project the final-position MLP output onto the reference-versus-contrast logit direction
\[
  \mathbf{d}_{c,\bar{c}}
  =
  W_U[:,c]-W_U[:,\bar{c}],
\]
where $W_U$ is the unembedding matrix.
Let $\widetilde{\mathbf{m}}^{(\ell)}(x)$ denote the layer-$\ell$ MLP output after the normalization scaling used for attribution.
We measure the anchor-dependent change in its contribution to the logit margin as
\begin{equation}
  \Delta \mathrm{DLA}^{(\ell)}
  =
  \left\langle
  \widetilde{\mathbf{m}}^{(\ell)}(x_{\mathrm{ref}})
  -
  \widetilde{\mathbf{m}}^{(\ell)}(x_{\mathrm{ctr}}),
  \,
  \mathbf{d}_{c,\bar{c}}
  \right\rangle.
  \label{eq:delta-dla}
\end{equation}
A positive value means that the MLP output favors $c$ over $\bar{c}$ more in the reference run than in the contrast run.
Because this comparison uses the unmodified runs, it captures all anchor-dependent information reaching the final-position MLPs, not only information arriving through the direct route; the value-injection experiment above provides the route-specific causal test.

\paragraph{\boxlabel{Result.}}
\underline{\emph{(i) Value injection.}}
Injecting the anchor signal recovers part of the reference--contrast logit margin, and almost all of this recovery passes through the MLPs: $R_{\mathrm{MLP}}=0.37$--$0.49$ against $R_{\mathrm{non\text{-}MLP}}\le 0.01$ across models (Fig.~\ref{fig:mlp_routed}).
What the final position reads from the anchor is therefore not yet answer-selective: without the MLPs' response, it barely moves the logits, and \textbf{it favors the year-appropriate object only after the final-position MLPs transform it.}

\underline{\emph{(ii) DLA.}}
$\Delta\mathrm{DLA}^{(\ell)}$ shows where the final-position MLPs become answer-selective.
In all four models, it stays near zero through the lower and middle layers and turns positive over a band of upper layers (Fig.~\ref{fig:mlp_readout}).
The MLPs in this band write an output favoring the reference-year object more in the reference run than in the contrast run, whereas earlier MLPs write little anchor-dependent preference.
\textbf{The anchor-dependent preference is thus written late, in the upper layers}, rather than accumulating gradually across the network.

Together, (i) and (ii) show how the direct route works: attention carries the anchor signal to the final position (\S\ref{sec:flow}), the final-position MLPs are necessary for its effect on the answer, and the preference for the year-appropriate object is written in the upper layers.
The recovery, however, is under half of the reference--contrast logit margin ($R_{\mathrm{total}}=R_{\mathrm{MLP}}+R_{\mathrm{non\text{-}MLP}}\le 0.49$), so this route does not determine the answer alone; how it combines with the subject-mediated route is the question of \S\ref{sec:two_routes_combination}.

\subsubsection{\routetitle{qacolor}{Route Combination}: Approximately Additive Effects with Query-Dependent Balance}
\label{sec:two_routes_combination}

\begin{figure}
  \centering
  \includegraphics[width=\linewidth]{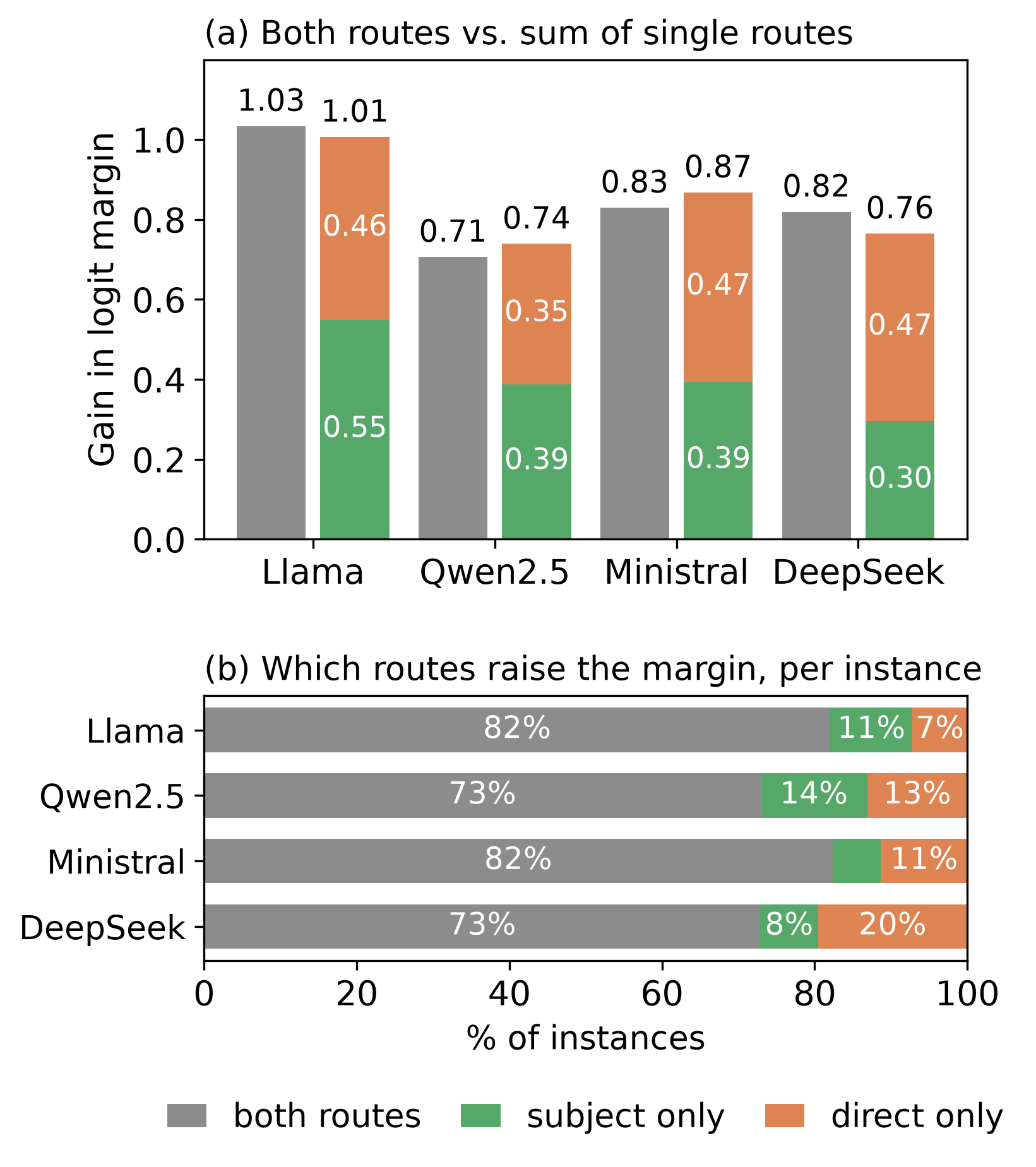}
  \caption{\takeaway{The two routes are parallel channels whose contributions add up, and most instances use both.}
    (a) Gain in the logit margin over the \emph{none} condition: both routes (left) vs.\ the two single-route gains, stacked (right).
  (b) Instances whose margin is raised by at least one route alone, by which routes raise it.}
  \label{fig:route_additivity}
\end{figure}
The preceding sections examine each route on its own.
Here we ask how their contributions combine: are their effects additive (i), and how do their relative contributions vary across queries (ii)?

\paragraph{\boxlabel{Method.}}
We control which positions can read the anchor.
At all layers, we block attention to the anchor span $\mathcal{A}$ from every later position except those we choose.
This gives four conditions: \emph{both} (the subject tokens and the final position can read the anchor), \emph{subject only}, \emph{direct only} (the final position), and \emph{none}.
In the unmodified run, \emph{full}, every position can read it.

For each condition we compute a margin: the logit of the correct object minus that of a competitor.
The competitor is the gold object of another year of the same group (the first such year in the dataset), and each object is scored by the maximum over the first tokens of its aliases.
The \emph{gain} of a condition is its margin minus the margin under \emph{none}.
We write $G_{\mathrm{full}}$, $G_{\mathrm{both}}$, $G_{\mathcal{S}}$, and $G_{\mathcal{L}}$ for the mean gains over instances under \emph{full}, \emph{both}, \emph{subject only}, and \emph{direct only}.
If the two routes combine additively, the two single-route gains should sum to the gain when both routes are available:
\begin{equation}
  G_{\mathcal{S}}+G_{\mathcal{L}}=G_{\mathrm{both}}.
  \label{eq:route-additivity}
\end{equation}

\paragraph{\boxlabel{Result.}}
\underline{\emph{(i) Additivity.}}
\textbf{The effects of the two routes are approximately additive.}
Each route raises the margin when available alone, and the two gains sum closely to the gain when both routes are available (Fig.~\ref{fig:route_additivity}a).
$G_{\mathcal{S}}+G_{\mathcal{L}}$ is within $5\%$ of $G_{\mathrm{both}}$ in Llama, Qwen, and Ministral, with no significant difference after Bonferroni correction, and $7\%$ below it in DeepSeek, a significant difference.
The gains also add up within individual instances (Appendix~\ref{app:route-additivity}).
These results suggest that the two routes contribute largely in parallel rather than one replacing the other.
Together, the two routes account for most of the anchor effect available in the unmodified model: $G_{\mathrm{both}}/G_{\mathrm{full}}$ is $0.69$--$0.90$ across models.
The remaining effect requires allowing positions other than the subject and final position to attend to the anchor.\footnote{This does not conflict with \S\ref{sec:flow}: these positions comprise more than the relation span, and they are blocked here at all layers rather than within a window.}

\underline{\emph{(ii) Variation across queries.}}
Both routes can independently support the prediction on most queries, but their relative contributions vary.
Counting a route as contributing when its single-route gain is above zero, $73$--$82\%$ of the instances with a positive gain from at least one route have positive gains from both (Fig.~\ref{fig:route_additivity}b; stricter criteria in Appendix~\ref{app:route-relation}).
Thus, for most queries, either route alone carries answer-relevant anchor information rather than the prediction depending exclusively on one route.
The balance between the two gains varies across instances (Appendix~\ref{app:route-scatter}), and the models agree on how it varies.
For subject--relation groups analyzed in two models, $G_{\mathcal{S}}-G_{\mathcal{L}}$ is positively correlated in all six model pairs (Spearman's $\rho=0.54$--$0.82$).
This cross-model agreement suggests that \textbf{the relative contribution of the two routes reflects properties of the queried fact rather than model-specific noise.}
One such property is the relation.
We measure the relative subject-route contribution as $G_{\mathcal{S}}/(G_{\mathcal{S}}+G_{\mathcal{L}})$.
For \emph{position held} (P39), this share is $64$--$75\%$ in all four models; for \emph{member of sports team} (P54), it is $20$--$46\%$, indicating a larger direct-route contribution (difference significant in every model, $p<0.001$).
The route-specific interventions show the same pattern.
Subject-state patching (\S\ref{sec:subject}) is stronger for P39, raising the logit of $c$ by $0.59$--$0.77$ of the reference--contrast difference (peak over layers), compared with $0.29$--$0.50$ for P54.
Conversely, value injection (\S\ref{sec:direct-mlp}) is stronger for P54, with $R_{\mathrm{MLP}}=0.43$--$0.68$, compared with $0.16$--$0.23$ for P39.
The relation therefore explains part of the variation in the balance between the two routes; which other query properties matter remains open.

\section{Related Work}
\label{sec:related}

\paragraph{Mechanistic studies of factual recall.}
Prior work traces how Transformers retrieve factual associations from subject--relation prompts~\citep{geva-etal-2023-dissecting}, localizing factual information and analyzing how attention, MLPs, and residual-stream representations contribute to predictions~\citep{meng2022locating,hernandez2024linearity}.
Assuming a fixed object per subject--relation pair, these studies leave open where competition among objects is resolved; we find it resolved both in the subject representation and at the final position, with the two contributing approximately additively.
Our question also differs from whether time is represented~\citep{gurnee2024language} or how context competes with parametric memory~\citep{yu2023characterizing,li2025taming}: we investigate how an explicit temporal anchor selects among time-dependent facts.

\paragraph{Temporal factual knowledge.}
TempLAMA and related work evaluate whether LM predictions change appropriately across timestamps or event times~\citep{dhingra2022time,son-oh-2023-time,chen2021a,wei2023menatqa}, assessing temporal knowledge at the output level.
\citet{park-etal-2025-time} go further and localize attention heads specialized for temporal knowledge; a single illustrative causal-tracing example there also hints that time is merged into the subject representation.
We establish this route across four models, show that the resulting state is largely entity-specific and selectively supports the year-appropriate object (\S\ref{sec:subject}), and identify a second, direct route and how the two combine (\S\ref{sec:flow}, \S\ref{sec:two_routes_combination}).
Concurrently, \citet{vanadrichem2026languagemodelsconsistentlyencode} characterize the mechanisms supporting an implicitly assumed current year, whereas we study an explicit anchor that selects among stored facts.

\paragraph{Context-dependent factual recall.}
Anchored factual recall relates to contextually mediated factual recall, where information beyond the subject and relation determines the answer.
Recent work shows that factual recall becomes less reliable when entities are specified through descriptive contexts~\citep{liu2026evaluatingcontextuallymediatedfactual}.
Our setting isolates an explicit conditioning expression that can be manipulated, enabling a causal analysis of how contextual information is integrated into factual prediction.

\section{Conclusion}
\label{sec:conclusion}

We introduced \emph{anchored factual recall}, where an explicit condition determines which object is valid for a subject--relation pair, and used temporal anchors to study how LMs implement this selection.
Across four open-weight models, we identified a \subjroute, in which the anchor resolves the final subject-token representation to the queried year and the relation redistributes last-subject readout across attention heads.
Cross-fitted ablation shows that the last-subject contribution of relation-sensitive heads is causally important for the model's originally correct next-token prediction.
We also identified a \dirroute, in which the final position accesses the anchor directly and final-position MLPs transform this signal into a preference for the time-appropriate object.
The two routes act as parallel channels: their contributions add up, most queries use both, and the balance between them varies across queries in the same way across models.
Anchored factual recall thus reframes the mechanistic study of factual knowledge as a problem of selecting among competing context-dependent facts, not merely retrieving a fixed association.

\section*{Limitations}
\label{sec:limit}
Our experiments cover one setting: English, one dataset (TempLAMA), one prompt format, and years as anchors.
We chose a single controlled setting so that reference and contrast prompts differ only in the anchor, and our claims are about this setting only.
We have not tested other prompt formats, other temporal expressions, such as dates or events, or non-temporal anchors, such as locations, so whether the two routes hold for anchored factual recall in general remains open.

The prompt format matters for the findings themselves.
The subject-mediated route requires the anchor to precede the subject, because under causal attention the subject can read only what comes before it.
We could not vary this order within English without a confound: placing the anchor after the subject while keeping the object last either changes the syntactic construction or yields sentences that are rare in natural English, so a change in route usage could not be separated from a reaction to unusual input.
Testing the role of word order therefore needs a language in which both orders are natural, and we state our claims for English only.

All four models have 7--8B parameters.
Smaller models answer too few time-varying facts correctly to pass our filter, and larger ones are costly to analyze: our interventions take about 150 times the forward passes of ordinary prediction per example.
Whether the two routes, and the way they combine, hold at other scales remains open.

\bibliography{custom}

\begin{thebibliography}{26}
\providecommand{\natexlab}[1]{#1}

\bibitem[{Chen et~al.(2021)Chen, Wang, and Wang}]{chen2021a}
Wenhu Chen, Xinyi Wang, and William~Yang Wang. 2021.
\newblock \href {https://openreview.net/forum?id=9-LSfSU74n-} {A dataset for answering time-sensitive questions}.
\newblock In \emph{Thirty-fifth Conference on Neural Information Processing Systems Datasets and Benchmarks Track (Round 2)}.

\bibitem[{Dai et~al.(2022)Dai, Dong, Hao, Sui, Chang, and Wei}]{dai2022knowledge}
Damai Dai, Li~Dong, Yaru Hao, Zhifang Sui, Baobao Chang, and Furu Wei. 2022.
\newblock Knowledge neurons in pretrained transformers.
\newblock In \emph{Proceedings of the 60th Annual Meeting of the Association for Computational Linguistics (Volume 1: Long Papers)}, pages 8493--8502.

\bibitem[{{DeepSeek-AI}(2025)}]{deepseekai2025deepseekr1incentivizingreasoningcapability}
{DeepSeek-AI}. 2025.
\newblock \href {https://arxiv.org/abs/2501.12948} {{DeepSeek-R1}: Incentivizing reasoning capability in {LLMs} via reinforcement learning}.
\newblock \emph{Preprint}, arXiv:2501.12948.

\bibitem[{Dhingra et~al.(2022)Dhingra, Cole, Eisenschlos, Gillick, Eisenstein, and Cohen}]{dhingra2022time}
Bhuwan Dhingra, Jeremy~R. Cole, Julian~Martin Eisenschlos, Daniel Gillick, Jacob Eisenstein, and William~W. Cohen. 2022.
\newblock \href {https://doi.org/10.1162/tacl_a_00459} {Time-aware language models as temporal knowledge bases}.
\newblock \emph{Transactions of the Association for Computational Linguistics}, 10:257--273.

\bibitem[{Geva et~al.(2023)Geva, Bastings, Filippova, and Globerson}]{geva-etal-2023-dissecting}
Mor Geva, Jasmijn Bastings, Katja Filippova, and Amir Globerson. 2023.
\newblock \href {https://doi.org/10.18653/v1/2023.emnlp-main.751} {Dissecting recall of factual associations in auto-regressive language models}.
\newblock In \emph{Proceedings of the 2023 Conference on Empirical Methods in Natural Language Processing}, pages 12216--12235, Singapore. Association for Computational Linguistics.

\bibitem[{Grattafiori et~al.(2024)Grattafiori, Dubey, Jauhri, Pandey, Kadian, Al-Dahle, Letman, Mathur, Schelten, Vaughan, Yang, Fan, Goyal, Hartshorn, Yang, Mitra, Sravankumar, Korenev, Hinsvark, Rao, Zhang, Rodriguez, Gregerson, Spataru, Roziere, Biron, Tang, Chern, Caucheteux, Nayak, Bi, Marra, McConnell, Keller, Touret, Wu, Wong, Ferrer, Nikolaidis, Allonsius, Song, Pintz, Livshits, Wyatt, Esiobu, Choudhary, Mahajan, Garcia-Olano, Perino, Hupkes, Lakomkin, AlBadawy, Lobanova, Dinan, Smith, Radenovic, Guzmán, Zhang, Synnaeve, Lee, Anderson, Thattai, Nail, Mialon, Pang, Cucurell, Nguyen, Korevaar, Xu, Touvron, Zarov, Ibarra, Kloumann, Misra, Evtimov, Zhang, Copet, Lee, Geffert, Vranes, Park, Mahadeokar, Shah, van~der Linde, Billock, Hong, Lee, Fu, Chi, Huang, Liu, Wang, Yu, Bitton, Spisak, Park, Rocca, Johnstun, Saxe, Jia, Alwala, Prasad, Upasani, Plawiak, Li, Heafield, Stone, El-Arini, Iyer, Malik, Chiu, Bhalla, Lakhotia, Rantala-Yeary, van~der Maaten, Chen, Tan, Jenkins, Martin, Madaan, Malo, Blecher,
  Landzaat, de~Oliveira, Muzzi, Pasupuleti, Singh, Paluri, Kardas, Tsimpoukelli, Oldham, Rita, Pavlova, Kambadur, Lewis, Si, Singh, Hassan, Goyal, Torabi, Bashlykov, Bogoychev, Chatterji, Zhang, Duchenne, Çelebi, Alrassy, Zhang, Li, Vasic, Weng, Bhargava, Dubal, Krishnan, Koura, Xu, He, Dong, Srinivasan, Ganapathy, Calderer, Cabral, Stojnic, Raileanu, Maheswari, Girdhar, Patel, Sauvestre, Polidoro, Sumbaly, Taylor, Silva, Hou, Wang, Hosseini, Chennabasappa, Singh, Bell, Kim, Edunov, Nie, Narang, Raparthy, Shen, Wan, Bhosale, Zhang, Vandenhende, Batra, Whitman, Sootla, Collot, Gururangan, Borodinsky, Herman, Fowler, Sheasha, Georgiou, Scialom, Speckbacher, Mihaylov, Xiao, Karn, Goswami, Gupta, Ramanathan, Kerkez, Gonguet, Do, Vogeti, Albiero, Petrovic, Chu, Xiong, Fu, Meers, Martinet, Wang, Wang, Tan, Xia, Xie, Jia, Wang, Goldschlag, Gaur, Babaei, Wen, Song, Zhang, Li, Mao, Coudert, Yan, Chen, Papakipos, Singh, Srivastava, Jain, Kelsey, Shajnfeld, Gangidi, Victoria, Goldstand, Menon, Sharma, Boesenberg,
  Baevski, Feinstein, Kallet, Sangani, Teo, Yunus, Lupu, Alvarado, Caples, Gu, Ho, Poulton, Ryan, Ramchandani, Dong, Franco, Goyal, Saraf, Chowdhury, Gabriel, Bharambe, Eisenman, Yazdan, James, Maurer, Leonhardi, Huang, Loyd, Paola, Paranjape, Liu, Wu, Ni, Hancock, Wasti, Spence, Stojkovic, Gamido, Montalvo, Parker, Burton, Mejia, Liu, Wang, Kim, Zhou, Hu, Chu, Cai, Tindal, Feichtenhofer, Gao, Civin, Beaty, Kreymer, Li, Adkins, Xu, Testuggine, David, Parikh, Liskovich, Foss, Wang, Le, Holland, Dowling, Jamil, Montgomery, Presani, Hahn, Wood, Le, Brinkman, Arcaute, Dunbar, Smothers, Sun, Kreuk, Tian, Kokkinos, Ozgenel, Caggioni, Kanayet, Seide, Florez, Schwarz, Badeer, Swee, Halpern, Herman, Sizov, Guangyi, Zhang, Lakshminarayanan, Inan, Shojanazeri, Zou, Wang, Zha, Habeeb, Rudolph, Suk, Aspegren, Goldman, Zhan, Damlaj, Molybog, Tufanov, Leontiadis, Veliche, Gat, Weissman, Geboski, Kohli, Lam, Asher, Gaya, Marcus, Tang, Chan, Zhen, Reizenstein, Teboul, Zhong, Jin, Yang, Cummings, Carvill, Shepard, McPhie,
  Torres, Ginsburg, Wang, Wu, U, Saxena, Khandelwal, Zand, Matosich, Veeraraghavan, Michelena, Li, Jagadeesh, Huang, Chawla, Huang, Chen, Garg, A, Silva, Bell, Zhang, Guo, Yu, Moshkovich, Wehrstedt, Khabsa, Avalani, Bhatt, Mankus, Hasson, Lennie, Reso, Groshev, Naumov, Lathi, Keneally, Liu, Seltzer, Valko, Restrepo, Patel, Vyatskov, Samvelyan, Clark, Macey, Wang, Hermoso, Metanat, Rastegari, Bansal, Santhanam, Parks, White, Bawa, Singhal, Egebo, Usunier, Mehta, Laptev, Dong, Cheng, Chernoguz, Hart, Salpekar, Kalinli, Kent, Parekh, Saab, Balaji, Rittner, Bontrager, Roux, Dollar, Zvyagina, Ratanchandani, Yuvraj, Liang, Alao, Rodriguez, Ayub, Murthy, Nayani, Mitra, Parthasarathy, Li, Hogan, Battey, Wang, Howes, Rinott, Mehta, Siby, Bondu, Datta, Chugh, Hunt, Dhillon, Sidorov, Pan, Mahajan, Verma, Yamamoto, Ramaswamy, Lindsay, Lindsay, Feng, Lin, Zha, Patil, Shankar, Zhang, Zhang, Wang, Agarwal, Sajuyigbe, Chintala, Max, Chen, Kehoe, Satterfield, Govindaprasad, Gupta, Deng, Cho, Virk, Subramanian, Choudhury,
  Goldman, Remez, Glaser, Best, Koehler, Robinson, Li, Zhang, Matthews, Chou, Shaked, Vontimitta, Ajayi, Montanez, Mohan, Kumar, Mangla, Ionescu, Poenaru, Mihailescu, Ivanov, Li, Wang, Jiang, Bouaziz, Constable, Tang, Wu, Wang, Wu, Gao, Kleinman, Chen, Hu, Jia, Qi, Li, Zhang, Zhang, Adi, Nam, Yu, Wang, Zhao, Hao, Qian, Li, He, Rait, DeVito, Rosnbrick, Wen, Yang, Zhao, and Ma}]{grattafiori2024llama3herdmodels}
Aaron Grattafiori, Abhimanyu Dubey, Abhinav Jauhri, Abhinav Pandey, Abhishek Kadian, Ahmad Al-Dahle, Aiesha Letman, Akhil Mathur, Alan Schelten, Alex Vaughan, Amy Yang, Angela Fan, Anirudh Goyal, Anthony Hartshorn, Aobo Yang, Archi Mitra, Archie Sravankumar, Artem Korenev, Arthur Hinsvark, Arun Rao, Aston Zhang, Aurelien Rodriguez, Austen Gregerson, Ava Spataru, Baptiste Roziere, Bethany Biron, Binh Tang, Bobbie Chern, Charlotte Caucheteux, Chaya Nayak, Chloe Bi, Chris Marra, Chris McConnell, Christian Keller, Christophe Touret, Chunyang Wu, Corinne Wong, Cristian~Canton Ferrer, Cyrus Nikolaidis, Damien Allonsius, Daniel Song, Danielle Pintz, Danny Livshits, Danny Wyatt, David Esiobu, Dhruv Choudhary, Dhruv Mahajan, Diego Garcia-Olano, Diego Perino, Dieuwke Hupkes, Egor Lakomkin, Ehab AlBadawy, Elina Lobanova, Emily Dinan, Eric~Michael Smith, Filip Radenovic, Francisco Guzmán, Frank Zhang, Gabriel Synnaeve, Gabrielle Lee, Georgia~Lewis Anderson, Govind Thattai, Graeme Nail, Gregoire Mialon, Guan Pang,
  Guillem Cucurell, Hailey Nguyen, Hannah Korevaar, Hu~Xu, Hugo Touvron, Iliyan Zarov, Imanol~Arrieta Ibarra, Isabel Kloumann, Ishan Misra, Ivan Evtimov, Jack Zhang, Jade Copet, Jaewon Lee, Jan Geffert, Jana Vranes, Jason Park, Jay Mahadeokar, Jeet Shah, Jelmer van~der Linde, Jennifer Billock, Jenny Hong, Jenya Lee, Jeremy Fu, Jianfeng Chi, Jianyu Huang, Jiawen Liu, Jie Wang, Jiecao Yu, Joanna Bitton, Joe Spisak, Jongsoo Park, Joseph Rocca, Joshua Johnstun, Joshua Saxe, Junteng Jia, Kalyan~Vasuden Alwala, Karthik Prasad, Kartikeya Upasani, Kate Plawiak, Ke~Li, Kenneth Heafield, Kevin Stone, Khalid El-Arini, Krithika Iyer, Kshitiz Malik, Kuenley Chiu, Kunal Bhalla, Kushal Lakhotia, Lauren Rantala-Yeary, Laurens van~der Maaten, Lawrence Chen, Liang Tan, Liz Jenkins, Louis Martin, Lovish Madaan, Lubo Malo, Lukas Blecher, Lukas Landzaat, Luke de~Oliveira, Madeline Muzzi, Mahesh Pasupuleti, Mannat Singh, Manohar Paluri, Marcin Kardas, Maria Tsimpoukelli, Mathew Oldham, Mathieu Rita, Maya Pavlova, Melanie Kambadur,
  Mike Lewis, Min Si, Mitesh~Kumar Singh, Mona Hassan, Naman Goyal, Narjes Torabi, Nikolay Bashlykov, Nikolay Bogoychev, Niladri Chatterji, Ning Zhang, Olivier Duchenne, Onur Çelebi, Patrick Alrassy, Pengchuan Zhang, Pengwei Li, Petar Vasic, Peter Weng, Prajjwal Bhargava, Pratik Dubal, Praveen Krishnan, Punit~Singh Koura, Puxin Xu, Qing He, Qingxiao Dong, Ragavan Srinivasan, Raj Ganapathy, Ramon Calderer, Ricardo~Silveira Cabral, Robert Stojnic, Roberta Raileanu, Rohan Maheswari, Rohit Girdhar, Rohit Patel, Romain Sauvestre, Ronnie Polidoro, Roshan Sumbaly, Ross Taylor, Ruan Silva, Rui Hou, Rui Wang, Saghar Hosseini, Sahana Chennabasappa, Sanjay Singh, Sean Bell, Seohyun~Sonia Kim, Sergey Edunov, Shaoliang Nie, Sharan Narang, Sharath Raparthy, Sheng Shen, Shengye Wan, Shruti Bhosale, Shun Zhang, Simon Vandenhende, Soumya Batra, Spencer Whitman, Sten Sootla, Stephane Collot, Suchin Gururangan, Sydney Borodinsky, Tamar Herman, Tara Fowler, Tarek Sheasha, Thomas Georgiou, Thomas Scialom, Tobias Speckbacher,
  Todor Mihaylov, Tong Xiao, Ujjwal Karn, Vedanuj Goswami, Vibhor Gupta, Vignesh Ramanathan, Viktor Kerkez, Vincent Gonguet, Virginie Do, Vish Vogeti, Vítor Albiero, Vladan Petrovic, Weiwei Chu, Wenhan Xiong, Wenyin Fu, Whitney Meers, Xavier Martinet, Xiaodong Wang, Xiaofang Wang, Xiaoqing~Ellen Tan, Xide Xia, Xinfeng Xie, Xuchao Jia, Xuewei Wang, Yaelle Goldschlag, Yashesh Gaur, Yasmine Babaei, Yi~Wen, Yiwen Song, Yuchen Zhang, Yue Li, Yuning Mao, Zacharie~Delpierre Coudert, Zheng Yan, Zhengxing Chen, Zoe Papakipos, Aaditya Singh, Aayushi Srivastava, Abha Jain, Adam Kelsey, Adam Shajnfeld, Adithya Gangidi, Adolfo Victoria, Ahuva Goldstand, Ajay Menon, Ajay Sharma, Alex Boesenberg, Alexei Baevski, Allie Feinstein, Amanda Kallet, Amit Sangani, Amos Teo, Anam Yunus, Andrei Lupu, Andres Alvarado, Andrew Caples, Andrew Gu, Andrew Ho, Andrew Poulton, Andrew Ryan, Ankit Ramchandani, Annie Dong, Annie Franco, Anuj Goyal, Aparajita Saraf, Arkabandhu Chowdhury, Ashley Gabriel, Ashwin Bharambe, Assaf Eisenman, Azadeh
  Yazdan, Beau James, Ben Maurer, Benjamin Leonhardi, Bernie Huang, Beth Loyd, Beto~De Paola, Bhargavi Paranjape, Bing Liu, Bo~Wu, Boyu Ni, Braden Hancock, Bram Wasti, Brandon Spence, Brani Stojkovic, Brian Gamido, Britt Montalvo, Carl Parker, Carly Burton, Catalina Mejia, Ce~Liu, Changhan Wang, Changkyu Kim, Chao Zhou, Chester Hu, Ching-Hsiang Chu, Chris Cai, Chris Tindal, Christoph Feichtenhofer, Cynthia Gao, Damon Civin, Dana Beaty, Daniel Kreymer, Daniel Li, David Adkins, David Xu, Davide Testuggine, Delia David, Devi Parikh, Diana Liskovich, Didem Foss, Dingkang Wang, Duc Le, Dustin Holland, Edward Dowling, Eissa Jamil, Elaine Montgomery, Eleonora Presani, Emily Hahn, Emily Wood, Eric-Tuan Le, Erik Brinkman, Esteban Arcaute, Evan Dunbar, Evan Smothers, Fei Sun, Felix Kreuk, Feng Tian, Filippos Kokkinos, Firat Ozgenel, Francesco Caggioni, Frank Kanayet, Frank Seide, Gabriela~Medina Florez, Gabriella Schwarz, Gada Badeer, Georgia Swee, Gil Halpern, Grant Herman, Grigory Sizov, Guangyi, Zhang, Guna
  Lakshminarayanan, Hakan Inan, Hamid Shojanazeri, Han Zou, Hannah Wang, Hanwen Zha, Haroun Habeeb, Harrison Rudolph, Helen Suk, Henry Aspegren, Hunter Goldman, Hongyuan Zhan, Ibrahim Damlaj, Igor Molybog, Igor Tufanov, Ilias Leontiadis, Irina-Elena Veliche, Itai Gat, Jake Weissman, James Geboski, James Kohli, Janice Lam, Japhet Asher, Jean-Baptiste Gaya, Jeff Marcus, Jeff Tang, Jennifer Chan, Jenny Zhen, Jeremy Reizenstein, Jeremy Teboul, Jessica Zhong, Jian Jin, Jingyi Yang, Joe Cummings, Jon Carvill, Jon Shepard, Jonathan McPhie, Jonathan Torres, Josh Ginsburg, Junjie Wang, Kai Wu, Kam~Hou U, Karan Saxena, Kartikay Khandelwal, Katayoun Zand, Kathy Matosich, Kaushik Veeraraghavan, Kelly Michelena, Keqian Li, Kiran Jagadeesh, Kun Huang, Kunal Chawla, Kyle Huang, Lailin Chen, Lakshya Garg, Lavender A, Leandro Silva, Lee Bell, Lei Zhang, Liangpeng Guo, Licheng Yu, Liron Moshkovich, Luca Wehrstedt, Madian Khabsa, Manav Avalani, Manish Bhatt, Martynas Mankus, Matan Hasson, Matthew Lennie, Matthias Reso, Maxim
  Groshev, Maxim Naumov, Maya Lathi, Meghan Keneally, Miao Liu, Michael~L. Seltzer, Michal Valko, Michelle Restrepo, Mihir Patel, Mik Vyatskov, Mikayel Samvelyan, Mike Clark, Mike Macey, Mike Wang, Miquel~Jubert Hermoso, Mo~Metanat, Mohammad Rastegari, Munish Bansal, Nandhini Santhanam, Natascha Parks, Natasha White, Navyata Bawa, Nayan Singhal, Nick Egebo, Nicolas Usunier, Nikhil Mehta, Nikolay~Pavlovich Laptev, Ning Dong, Norman Cheng, Oleg Chernoguz, Olivia Hart, Omkar Salpekar, Ozlem Kalinli, Parkin Kent, Parth Parekh, Paul Saab, Pavan Balaji, Pedro Rittner, Philip Bontrager, Pierre Roux, Piotr Dollar, Polina Zvyagina, Prashant Ratanchandani, Pritish Yuvraj, Qian Liang, Rachad Alao, Rachel Rodriguez, Rafi Ayub, Raghotham Murthy, Raghu Nayani, Rahul Mitra, Rangaprabhu Parthasarathy, Raymond Li, Rebekkah Hogan, Robin Battey, Rocky Wang, Russ Howes, Ruty Rinott, Sachin Mehta, Sachin Siby, Sai~Jayesh Bondu, Samyak Datta, Sara Chugh, Sara Hunt, Sargun Dhillon, Sasha Sidorov, Satadru Pan, Saurabh Mahajan,
  Saurabh Verma, Seiji Yamamoto, Sharadh Ramaswamy, Shaun Lindsay, Shaun Lindsay, Sheng Feng, Shenghao Lin, Shengxin~Cindy Zha, Shishir Patil, Shiva Shankar, Shuqiang Zhang, Shuqiang Zhang, Sinong Wang, Sneha Agarwal, Soji Sajuyigbe, Soumith Chintala, Stephanie Max, Stephen Chen, Steve Kehoe, Steve Satterfield, Sudarshan Govindaprasad, Sumit Gupta, Summer Deng, Sungmin Cho, Sunny Virk, Suraj Subramanian, Sy~Choudhury, Sydney Goldman, Tal Remez, Tamar Glaser, Tamara Best, Thilo Koehler, Thomas Robinson, Tianhe Li, Tianjun Zhang, Tim Matthews, Timothy Chou, Tzook Shaked, Varun Vontimitta, Victoria Ajayi, Victoria Montanez, Vijai Mohan, Vinay~Satish Kumar, Vishal Mangla, Vlad Ionescu, Vlad Poenaru, Vlad~Tiberiu Mihailescu, Vladimir Ivanov, Wei Li, Wenchen Wang, Wenwen Jiang, Wes Bouaziz, Will Constable, Xiaocheng Tang, Xiaojian Wu, Xiaolan Wang, Xilun Wu, Xinbo Gao, Yaniv Kleinman, Yanjun Chen, Ye~Hu, Ye~Jia, Ye~Qi, Yenda Li, Yilin Zhang, Ying Zhang, Yossi Adi, Youngjin Nam, Yu, Wang, Yu~Zhao, Yuchen Hao, Yundi
  Qian, Yunlu Li, Yuzi He, Zach Rait, Zachary DeVito, Zef Rosnbrick, Zhaoduo Wen, Zhenyu Yang, Zhiwei Zhao, and Zhiyu Ma. 2024.
\newblock \href {https://arxiv.org/abs/2407.21783} {The {Llama} 3 herd of models}.
\newblock \emph{Preprint}, arXiv:2407.21783.

\bibitem[{Gurnee and Tegmark(2024)}]{gurnee2024language}
Wes Gurnee and Max Tegmark. 2024.
\newblock \href {https://openreview.net/forum?id=jE8xbmvFin} {Language models represent space and time}.
\newblock In \emph{The Twelfth International Conference on Learning Representations}.

\bibitem[{Herel et~al.(2025)Herel, Bartek, Jirak, and Mikolov}]{herel2025timeawarenesslargelanguage}
David Herel, Vojtech Bartek, Jiri Jirak, and Tomas Mikolov. 2025.
\newblock \href {https://arxiv.org/abs/2409.13338} {Time awareness in large language models: Benchmarking fact recall across time}.
\newblock \emph{Preprint}, arXiv:2409.13338.

\bibitem[{Hernandez et~al.(2024)Hernandez, Sharma, Haklay, Meng, Wattenberg, Andreas, Belinkov, and Bau}]{hernandez2024linearity}
Evan Hernandez, Arnab~Sen Sharma, Tal Haklay, Kevin Meng, Martin Wattenberg, Jacob Andreas, Yonatan Belinkov, and David Bau. 2024.
\newblock Linearity of relation decoding in transformer language models.
\newblock In \emph{Proceedings of the 2024 International Conference on Learning Representations}.

\bibitem[{Li et~al.(2025)Li, Chen, and Tong}]{li2025taming}
Gaotang Li, Yuzhong Chen, and Hanghang Tong. 2025.
\newblock \href {https://openreview.net/forum?id=0cEZyhHEks} {Taming knowledge conflicts in language models}.
\newblock In \emph{Forty-second International Conference on Machine Learning}.

\bibitem[{Liu et~al.(2026{\natexlab{a}})Liu, Khandelwal, Subramanian, Jouault, Rastogi, Sadé, Jeffares, Jiang, Cahill, Gavaudan, Sablayrolles, Héliou, You, Ehrenberg, Lo, Eliseev, Calvi, Sooriyarachchi, Bout, Rozière, Monicault, Lanfranchi, Barreau, Courtot, Grattarola, Dabert, de~las Casas, Chane-Sane, Ahmed, Berrada, Ecrepont, Guinet, Novikov, Kunsch, Lample, Martin, Gupta, Ludziejewski, Rute, Studnia, Amar, Delas, Roberts, Yadav, Chandu, Jain, Aitchison, Fainsin, Blier, Zhao, Martin, Saulnier, Gao, Buyl, Jennings, Pellat, Prins, Poirée, Guillaumin, Dinot, Futeral, Darrin, Augustin, Chiquier, Schimpf, Grinsztajn, Gupta, Raghuraman, Bousquet, Duchenne, Wang, von Platen, Jacob, Wambergue, Kurylowicz, Muddireddy, Chagniot, Stock, Agrawal, Torroba, Sauvestre, Soletskyi, Menneer, Vaze, Barry, Gandhi, Waghjale, Gandhi, Ghosh, Mishra, Aithal, Antoniak, Scao, Cachet, Sorg, Lavril, Saada, Chabal, Foubert, Robert, Wang, Lawson, Bewley, Bewley, Edwards, Jamil, Tomasini, Nemychnikova, Phung, Maladière, Richard,
  Bouaziz, Li, Marshall, Li, Yang, Ouahidi, Wang, Tang, and Ramzi}]{liu2026ministral3}
Alexander~H. Liu, Kartik Khandelwal, Sandeep Subramanian, Victor Jouault, Abhinav Rastogi, Adrien Sadé, Alan Jeffares, Albert Jiang, Alexandre Cahill, Alexandre Gavaudan, Alexandre Sablayrolles, Amélie Héliou, Amos You, Andy Ehrenberg, Andy Lo, Anton Eliseev, Antonia Calvi, Avinash Sooriyarachchi, Baptiste Bout, Baptiste Rozière, Baudouin~De Monicault, Clémence Lanfranchi, Corentin Barreau, Cyprien Courtot, Daniele Grattarola, Darius Dabert, Diego de~las Casas, Elliot Chane-Sane, Faruk Ahmed, Gabrielle Berrada, Gaëtan Ecrepont, Gauthier Guinet, Georgii Novikov, Guillaume Kunsch, Guillaume Lample, Guillaume Martin, Gunshi Gupta, Jan Ludziejewski, Jason Rute, Joachim Studnia, Jonas Amar, Joséphine Delas, Josselin~Somerville Roberts, Karmesh Yadav, Khyathi Chandu, Kush Jain, Laurence Aitchison, Laurent Fainsin, Léonard Blier, Lingxiao Zhao, Louis Martin, Lucile Saulnier, Luyu Gao, Maarten Buyl, Margaret Jennings, Marie Pellat, Mark Prins, Mathieu Poirée, Mathilde Guillaumin, Matthieu Dinot, Matthieu
  Futeral, Maxime Darrin, Maximilian Augustin, Mia Chiquier, Michel Schimpf, Nathan Grinsztajn, Neha Gupta, Nikhil Raghuraman, Olivier Bousquet, Olivier Duchenne, Patricia Wang, Patrick von Platen, Paul Jacob, Paul Wambergue, Paula Kurylowicz, Pavankumar~Reddy Muddireddy, Philomène Chagniot, Pierre Stock, Pravesh Agrawal, Quentin Torroba, Romain Sauvestre, Roman Soletskyi, Rupert Menneer, Sagar Vaze, Samuel Barry, Sanchit Gandhi, Siddhant Waghjale, Siddharth Gandhi, Soham Ghosh, Srijan Mishra, Sumukh Aithal, Szymon Antoniak, Teven~Le Scao, Théo Cachet, Theo~Simon Sorg, Thibaut Lavril, Thiziri~Nait Saada, Thomas Chabal, Thomas Foubert, Thomas Robert, Thomas Wang, Tim Lawson, Tom Bewley, Tom Bewley, Tom Edwards, Umar Jamil, Umberto Tomasini, Valeriia Nemychnikova, Van Phung, Vincent Maladière, Virgile Richard, Wassim Bouaziz, Wen-Ding Li, William Marshall, Xinghui Li, Xinyu Yang, Yassine~El Ouahidi, Yihan Wang, Yunhao Tang, and Zaccharie Ramzi. 2026{\natexlab{a}}.
\newblock \href {https://arxiv.org/abs/2601.08584} {Ministral 3}.
\newblock \emph{Preprint}, arXiv:2601.08584.

\bibitem[{Liu et~al.(2026{\natexlab{b}})Liu, Xiong, and Schütze}]{liu2026evaluatingcontextuallymediatedfactual}
Yihong Liu, Bingyu Xiong, and Hinrich Schütze. 2026{\natexlab{b}}.
\newblock \href {https://arxiv.org/abs/2601.12555} {Evaluating contextually mediated factual recall in multilingual large language models}.
\newblock \emph{Preprint}, arXiv:2601.12555.

\bibitem[{Meng et~al.(2022)Meng, Bau, Andonian, and Belinkov}]{meng2022locating}
Kevin Meng, David Bau, Alex Andonian, and Yonatan Belinkov. 2022.
\newblock Locating and editing factual associations in {GPT}.
\newblock \emph{Advances in neural information processing systems}, 35:17359--17372.

\bibitem[{Niwa et~al.(2025)Niwa, Kaneko, and Inui}]{niwa2025rectifying}
Ayana Niwa, Masahiro Kaneko, and Kentaro Inui. 2025.
\newblock Rectifying belief space via unlearning to harness {LLMs}' reasoning.
\newblock In \emph{Findings of the Association for Computational Linguistics: ACL 2025}, pages 25060--25075.

\bibitem[{nostalgebraist(2020)}]{nostalgebraist2020logitlens}
nostalgebraist. 2020.
\newblock interpreting {GPT}: the logit lens.
\newblock \url{https://www.lesswrong.com/posts/AcKRB8wDpdaN6v6ru/interpreting-gpt-the-logit-lens}.

\bibitem[{Park et~al.(2025)Park, Yoon, Park, Jeong, and Kang}]{park-etal-2025-time}
Yein Park, Chanwoong Yoon, Jungwoo Park, Minbyul Jeong, and Jaewoo Kang. 2025.
\newblock \href {https://doi.org/10.18653/v1/2025.acl-long.812} {Does time have its place? temporal heads: Where language models recall time-specific information}.
\newblock In \emph{Proceedings of the 63rd Annual Meeting of the Association for Computational Linguistics (Volume 1: Long Papers)}, pages 16616--16643, Vienna, Austria. Association for Computational Linguistics.

\bibitem[{Petroni et~al.(2019)Petroni, Rockt{\"a}schel, Riedel, Lewis, Bakhtin, Wu, and Miller}]{petroni2019language}
Fabio Petroni, Tim Rockt{\"a}schel, Sebastian Riedel, Patrick Lewis, Anton Bakhtin, Yuxiang Wu, and Alexander Miller. 2019.
\newblock Language models as knowledge bases?
\newblock In \emph{Proceedings of the 2019 conference on empirical methods in natural language processing and the 9th international joint conference on natural language processing (EMNLP-IJCNLP)}, pages 2463--2473.

\bibitem[{Qwen et~al.(2025)Qwen, :, Yang, Yang, Zhang, Hui, Zheng, Yu, Li, Liu, Huang, Wei, Lin, Yang, Tu, Zhang, Yang, Yang, Zhou, Lin, Dang, Lu, Bao, Yang, Yu, Li, Xue, Zhang, Zhu, Men, Lin, Li, Tang, Xia, Ren, Ren, Fan, Su, Zhang, Wan, Liu, Cui, Zhang, and Qiu}]{qwen2.5}
Qwen, :, An~Yang, Baosong Yang, Beichen Zhang, Binyuan Hui, Bo~Zheng, Bowen Yu, Chengyuan Li, Dayiheng Liu, Fei Huang, Haoran Wei, Huan Lin, Jian Yang, Jianhong Tu, Jianwei Zhang, Jianxin Yang, Jiaxi Yang, Jingren Zhou, Junyang Lin, Kai Dang, Keming Lu, Keqin Bao, Kexin Yang, Le~Yu, Mei Li, Mingfeng Xue, Pei Zhang, Qin Zhu, Rui Men, Runji Lin, Tianhao Li, Tianyi Tang, Tingyu Xia, Xingzhang Ren, Xuancheng Ren, Yang Fan, Yang Su, Yichang Zhang, Yu~Wan, Yuqiong Liu, Zeyu Cui, Zhenru Zhang, and Zihan Qiu. 2025.
\newblock \href {https://arxiv.org/abs/2412.15115} {Qwen2.5 technical report}.
\newblock \emph{Preprint}, arXiv:2412.15115.

\bibitem[{Roberts et~al.(2020)Roberts, Raffel, and Shazeer}]{roberts2020much}
Adam Roberts, Colin Raffel, and Noam Shazeer. 2020.
\newblock How much knowledge can you pack into the parameters of a language model?
\newblock In \emph{Proceedings of the 2020 conference on empirical methods in natural language processing (EMNLP)}, pages 5418--5426.

\bibitem[{Son and Oh(2023)}]{son-oh-2023-time}
Jungbin Son and Alice Oh. 2023.
\newblock \href {https://doi.org/10.18653/v1/2023.findings-emnlp.6} {Time-aware representation learning for time-sensitive question answering}.
\newblock In \emph{Findings of the Association for Computational Linguistics: EMNLP 2023}, pages 70--77, Singapore. Association for Computational Linguistics.

\bibitem[{Tan et~al.(2023)Tan, Ng, and Bing}]{tan-etal-2023-towards}
Qingyu Tan, Hwee~Tou Ng, and Lidong Bing. 2023.
\newblock \href {https://doi.org/10.18653/v1/2023.acl-long.828} {Towards benchmarking and improving the temporal reasoning capability of large language models}.
\newblock In \emph{Proceedings of the 61st Annual Meeting of the Association for Computational Linguistics (Volume 1: Long Papers)}, pages 14820--14835, Toronto, Canada. Association for Computational Linguistics.

\bibitem[{van Adrichem et~al.(2026)van Adrichem, Bhaskar, Yang, Potts, and Huang}]{vanadrichem2026languagemodelsconsistentlyencode}
Suze van Adrichem, Aditi Bhaskar, Diyi Yang, Christopher Potts, and Jing Huang. 2026.
\newblock \href {https://arxiv.org/abs/2608.15507} {Do language models consistently encode the current year?}
\newblock \emph{Preprint}, arXiv:2608.15507.

\bibitem[{Wang et~al.(2023)Wang, Variengien, Conmy, Shlegeris, and Steinhardt}]{wang2023interpretability}
Kevin Wang, Alexandre Variengien, Arthur Conmy, Buck Shlegeris, and Jacob Steinhardt. 2023.
\newblock Interpretability in the wild: a circuit for indirect object identification in {GPT-2} small.
\newblock In \emph{The International Conference on Learning Representations}.

\bibitem[{Wei et~al.(2023)Wei, Su, Ma, Yu, Lei, Zhang, Zhao, and Liu}]{wei2023menatqa}
Yifan Wei, Yisong Su, Huanhuan Ma, Xiaoyan Yu, Fangyu Lei, Yuanzhe Zhang, Jun Zhao, and Kang Liu. 2023.
\newblock {MenatQA}: A new dataset for testing the temporal comprehension and reasoning abilities of large language models.
\newblock In \emph{Findings of the Association for Computational Linguistics: EMNLP 2023}, pages 1434--1447.

\bibitem[{Yu et~al.(2023)Yu, Merullo, and Pavlick}]{yu2023characterizing}
Qinan Yu, Jack Merullo, and Ellie Pavlick. 2023.
\newblock Characterizing mechanisms for factual recall in language models.
\newblock In \emph{Proceedings of the 2023 Conference on Empirical Methods in Natural Language Processing}, pages 9924--9959.

\bibitem[{Zhao et~al.(2024)Zhao, Brumbaugh, Wang, Hajishirzi, and Smith}]{zhao-etal-2024-set}
Bowen Zhao, Zander Brumbaugh, Yizhong Wang, Hannaneh Hajishirzi, and Noah Smith. 2024.
\newblock \href {https://doi.org/10.18653/v1/2024.findings-acl.892} {Set the clock: Temporal alignment of pretrained language models}.
\newblock In \emph{Findings of the Association for Computational Linguistics: ACL 2024}, pages 15015--15040, Bangkok, Thailand. Association for Computational Linguistics.

\end{thebibliography}

\appendix
\section{Experimental Details}
\subsection{Dataset}
We evaluate on the TempLAMA test set, which contains 34,962 instances
and is publicly available at \url{https://github.com/google-research/language/tree/master/language/templama}.

\subsection{Implementation Details}
\label{app:impl}

\paragraph{Reference--contrast construction.}
Most analyses use temporal reference--contrast pairs that differ only in the anchor.

The relation-readout analysis additionally uses two swaps.
A length-matched \emph{relation swap} holds the anchor, subject, and demonstrations fixed while replacing the complete relation expression $\mathcal{R}\cup\mathcal{L}$ with that of another relation.
Each relation has one fixed replacement with the same number of tokens (e.g., ``\textit{played for}'' $\rightarrow$ ``\textit{was attending}'', ``\textit{held the position of}'' $\rightarrow$ ``\textit{was the chair of}''), so that all token positions stay aligned.
A \emph{time swap} holds the subject, relation, and demonstrations fixed while replacing only the anchor year.

These swaps are used for the descriptive readout analysis and for cross-fitted head selection.
The causal ablation itself is evaluated on the original, unmodified, correctly answered prompt.

\paragraph{Few-shot prompting.}
All prompts use a fixed $k=5$ shot format.
Demonstrations are sampled at random (fixed seed) from instances whose year, subject, and relation all differ from those of the target, so that no demonstration leaks the target answer or its exact query pattern.
Demonstrations are drawn independently per target instance.

\paragraph{Intervention scope.}
Attention knockout is applied over a centered window of \textbf{9} adjacent layers: an edge centered at layer $\ell$ is removed at that layer and its four neighbors on each side, following the windowed-knockout practice of \citet{geva-etal-2023-dissecting}.
The reported curve is indexed by the window center.
Subject-state activation patching is applied at a single layer at a time, yielding one value per layer.
Value injection is applied at all layers simultaneously, each layer receiving its own reference-minus-contrast difference, and MLP pinning to all final-position MLPs; the recovery decomposition therefore yields one value per pair.
The route-isolation analysis (\S\ref{sec:two_routes_combination}) and the anchor blocking of the shared-year-offset control (\S\ref{sec:subject-additive-null}) apply attention knockout at all layers.
The relation-sensitive-head ablation is the exception: it jointly intervenes on a cross-fitted set of layer--head pairs across multiple layers, as detailed below.

\paragraph{Cross-fitted relation-sensitive-head ablation.}
We partition subject--relation groups into five folds.
For each held-out fold, head selection and control construction use only the examples in the remaining four discovery folds.
We average the relation-sensitivity score $s_{\ell h}$ over the discovery examples and rank all finite layer--head scores across the analyzed layers.
We target the top $10\%$ of layer--head pairs, with a minimum of eight selected heads.
When at least eight positively scored candidates are available, selection is restricted to positive scores.
To preserve a disjoint same-layer pool for the random control, at most half of the heads in any layer may be selected.
Time-sensitive heads are selected using the reversed score and are matched to the relation-sensitive set in their per-layer head counts.
For each fold, we additionally construct 20 contribution-matched random head sets.
Each random set contains the same number of heads in each layer as the relation-sensitive set and excludes the selected relation-sensitive heads.
For each selected head, a random match is sampled without replacement from a pool of up to four unselected heads in the same layer whose mean unablated contribution norms on the discovery folds are closest to that of the selected head.
The reported matched-random effect is the mean over these 20 sets.
Evaluation is performed on the original prompts in the held-out fold.
For every selected layer--head pair, we recompute its current contribution from the final subject token $s^\star$ to the final query position during the modified forward pass, $\mathbf{c}_{\ell h}=\alpha_{\ell h}\,W_O^{(\ell,h)}\,\mathbf{V}^{(\ell,h)}(s^\star)$ (attention weight, output projection, and value vector of head $h$ in layer $\ell$; Eq.~\ref{eq:relation-head-contribution}), and subtract it from the attention output.
All selected pairs are ablated jointly in the same forward pass, so interventions in later layers operate on the state produced by the earlier interventions.
We report the top-1 flip rate defined in \S\ref{sec:relation-readout}.
Uncertainty is estimated with 2,000 percentile bootstrap resamples of subject--relation groups, yielding 95\% confidence intervals.
The same cluster bootstrap is applied to the condition means and to the paired differences between relation-sensitive and contribution-matched random ablation and between relation-sensitive and time-sensitive ablation.

\paragraph{Computational resources.}

All experiments were conducted on a single NVIDIA A40 GPU (48GB).

\begin{figure*}[t]
  \centering
  \includegraphics[width=\linewidth]{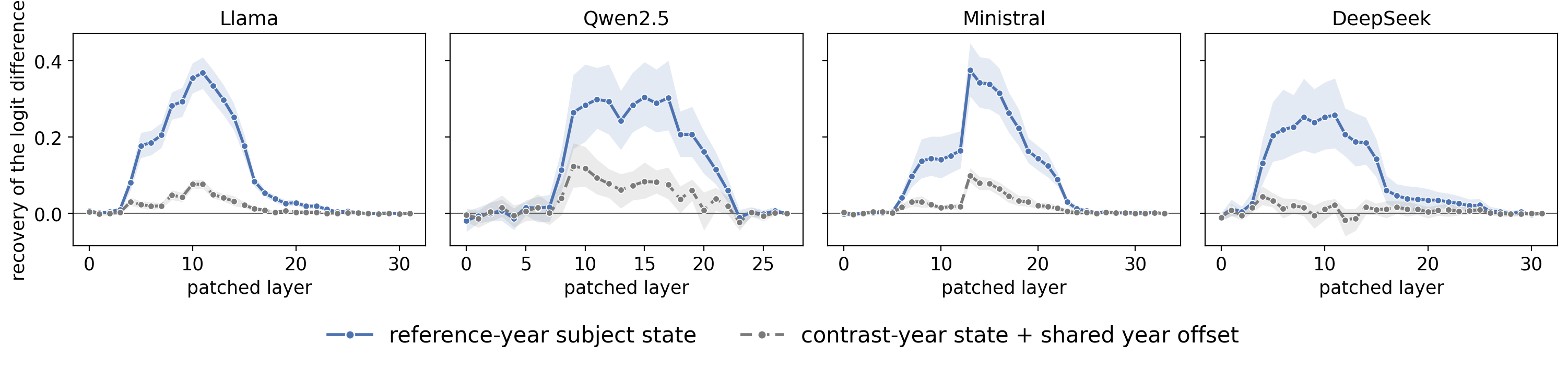}
  \caption{ \takeaway{The effect of the anchored subject state is not a shared year offset.} Recovery of the logit difference between $c$ and $\bar{c}$ when the final subject-token state of the contrast run is set to the reference-year state, or to the contrast-year state plus the mean reference-minus-contrast difference of other subjects with the same pair of years. No query position can attend to the anchor in either condition. Shaded bands are 95\% bootstrap intervals over subject--relation groups. }
  \label{fig:additive_null}
\end{figure*}
\begin{figure*}[t]
  \centering
  \includegraphics[width=\linewidth]{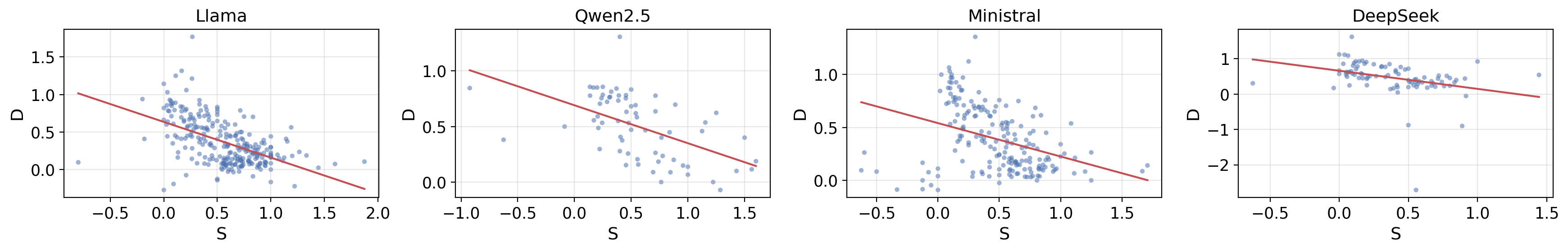}
  \caption{\takeaway{Route usage varies across instances.}
  Subject-mediated recovery ($S_i$) vs.\ direct-route recovery ($D_i$) per reference--contrast comparison.}
  \label{fig:corr}
\end{figure*}
\section{Additional Analyses}
\subsection{Control: The Effect Is Not a Shared Year Offset} \label{sec:subject-additive-null}

\paragraph{\boxlabel{Method.}}
Suppose, for each pair of years, the anchor adds an entity-independent vector to the subject representation; we call this a \emph{shared year offset}.
The reference-year state would then be reproduced by adding the offset to the contrast-year state:
\[
  \mathbf{h}_{\mathcal{S},i}^{(\ell)}(t)
  \;\approx\;
  \mathbf{h}_{\mathcal{S},i}^{(\ell)}(\bar t) + \boldsymbol{\tau}^{(\ell)}(t,\bar t),
\]
where $t$ and $\bar t$ are the reference and contrast years of comparison $i$ and $\boldsymbol{\tau}^{(\ell)}(t,\bar t)$ does not depend on the entity.

For each comparison, we estimate $\boldsymbol{\tau}^{(\ell)}(t,\bar t)$ as the mean difference between the reference-year and contrast-year final subject-token states of \emph{other} subjects with the same pair of years.
Subjects are assigned to five folds and the estimate for a comparison uses only subjects from the other folds, with states first averaged within subject.
We then run the contrast prompt and set the final subject-token state at layer $\ell$ either to the true reference-year state or to the contrast-year state plus the estimated offset.
In both cases, attention to the anchor from the subject, the final position, and the remaining query positions is blocked in all layers, so that neither condition can recover the answer by reading the anchor elsewhere.
We report the recovery of the logit difference between $c$ and $\bar c$, measured relative to the same run with the contrast-year state and normalized by the difference between the unmodified reference and contrast runs.

\paragraph{\boxlabel{Result.}}
A shared year offset reproduces only a small part of the effect of the reference-year state (Fig.~\ref{fig:additive_null}).
At the layer where the two differ most, the true state recovers $0.37$, $0.30$, $0.38$, and $0.24$ of the logit difference in Llama, Qwen, Ministral, and DeepSeek, whereas the offset recovers $0.08$, $0.08$, $0.10$, and $-0.01$; the difference is significant in every model.
In each model, the offset thus accounts for at most about a quarter of the effect ($21\%$, $25\%$, $26\%$, and $0\%$).
The true state is most effective at the layers where subject-state patching peaks (Fig.~\ref{fig:single_year}).
The selective effect observed under patching therefore primarily reflects a state specific to the entity, rather than a year component common to all entities.

\subsection{Route Usage}

\subsubsection{Additivity of the two routes}
\label{app:route-additivity}
The two routes together account for $G_{\mathrm{both}}/G_{\mathrm{full}}=0.70$, $0.79$, $0.90$, and $0.69$ of the anchor's effect in Llama, Qwen, Ministral, and DeepSeek.

We test $G_{\mathcal{S}}+G_{\mathcal{L}}=G_{\mathrm{both}}$ (\S\ref{sec:two_routes_combination}) with a bootstrap over subject--relation groups.
The uncorrected $p$-values are $0.049$ for Llama, $0.39$ for Qwen, $0.10$ for Ministral, and $<0.001$ for DeepSeek; with a Bonferroni correction for the four models ($\alpha=0.0125$), only DeepSeek differs significantly.

Within instances, the deviation $G_{\mathcal{S},i}+G_{\mathcal{L},i}-G_{\mathrm{both},i}$ has median $-0.01$, $0.00$, $0.00$, and $-0.02$ and interquartile range $[-0.10, 0.04]$, $[-0.05, 0.05]$, $[-0.03, 0.07]$, and $[-0.14, 0.02]$ for Llama, Qwen, Ministral, and DeepSeek.
Its absolute value has a median of $0.05$--$0.07$ and is below $0.25$ in $85\%$, $91\%$, $82\%$, and $81\%$ of the instances.
For comparison, $G_{\mathrm{both},i}$ has a median of $0.34$--$0.81$ and interquartile range $[0.18, 1.67]$, $[0.00, 0.96]$, $[0.02, 1.30]$, and $[0.00, 1.37]$.

\subsubsection{Route usage across instances}
\label{app:route-scatter}
We compare, for each reference--contrast comparison $i$, how much each route contributes according to the route-specific interventions of \S\ref{sec:subject} and \S\ref{sec:direct-mlp}.
For the subject-mediated route, $S_i$ is the peak increase, over layers, in the logit of $c$ under subject-state patching, normalized by the difference in that logit between the reference and contrast runs.
For the direct route, $D_i=R_{\mathrm{MLP},i}$ is the MLP-mediated recovery of value injection.
The two scores are negatively correlated across instances in all four models (Fig.~\ref{fig:corr}): Spearman's $\rho=-0.55$ for Llama, $-0.62$ for Qwen, $-0.42$ for Ministral, and $-0.51$ for DeepSeek, all $p<10^{-6}$.
The share of the two routes thus varies across instances.
This does not conflict with additivity: $S_i$ and $D_i$ are both fractions of the same reference--contrast difference, so when the two contributions add up to a roughly fixed total, a larger share of one implies a smaller share of the other.
This describes how the usage of the two routes varies across instances; it does not imply that one route increases its contribution when the other is weakened, which we test next.

\subsubsection{Does weakening one route strengthen the other?}
\label{app:route-compensation}
We also tested the interaction between the routes by intervention.
We replace the states of the subject tokens, at all layers, with the states they have when no position can read the anchor, and measure the MLP-mediated recovery $R_{\mathrm{MLP}}$ of value injection (\S\ref{sec:direct-mlp}).
$R_{\mathrm{MLP}}$ is the recovery with the final-position MLPs free minus the recovery with them pinned to their contrast-run outputs; it is therefore also defined without value injection, where it measures the part of the replacement's own effect that passes through these MLPs.
The replacement raises $R_{\mathrm{MLP}}$, but the same replacement \emph{without} value injection raises it by the same amount ($0.17$--$0.30$ across models), so the increase is caused by the replacement itself, not by a stronger direct route.
After subtracting this control, the change in the direct route's contribution is $-0.015$ $[-0.031, 0.002]$ in Llama, $-0.045$ $[-0.103, 0.003]$ in Qwen, $0.010$ $[-0.012, 0.033]$ in Ministral, and $0.004$ $[-0.052, 0.092]$ in DeepSeek (95\% bootstrap intervals over subject--relation groups).
Weakening the subject-mediated route therefore does not strengthen the direct route, consistent with the additivity observed under route isolation (\S\ref{sec:two_routes_combination}).

\subsubsection{Route usage by instance and by relation}
\label{app:route-relation}
Fig.~\ref{fig:route_additivity}b counts a route as contributing when its gain is above zero.
If we require a gain above $0.1$--$0.3$ instead, both routes raise the margin in $42$--$66\%$ of the instances raised by at least one route.
In every model, instances raised by both routes are more frequent than they would be if the two routes acted on instances independently of each other (by $4$--$9$ percentage points at a threshold of zero; all 95\% intervals above zero).

The share of the subject-mediated route, $G_{\mathcal{S}}/(G_{\mathcal{S}}+G_{\mathcal{L}})$, differs between \emph{position held} (P39) and \emph{member of sports team} (P54) by $0.29$--$0.44$ across models (all $p<0.001$; bootstrap over subjects, Bonferroni-corrected for the four models).
Among the relations with fewer instances, the share is $14$--$34\%$ for \emph{owned by} (P127; three models, below $50\%$ in each with a 95\% bootstrap interval over instances) and about $50\%$ for \emph{employer} (P108).

\end{document}